\documentclass{article} % For LaTeX2e
\usepackage{iclr2021_conference,times}

\usepackage{amsmath,amsfonts,bm}

\def\eqref#1{equation~\ref{#1}}
\def\1{\bm{1}}

\DeclareMathAlphabet{\mathsfit}{\encodingdefault}{\sfdefault}{m}{sl}
\SetMathAlphabet{\mathsfit}{bold}{\encodingdefault}{\sfdefault}{bx}{n}

\usepackage{cite}
\usepackage{hyperref,cleveref}
\usepackage{url}
\usepackage[ruled,vlined]{algorithm2e}
\usepackage{graphicx}
\usepackage{wrapfig}

\title{Certifiably-Robust Federated Adversarial Learning via Randomized Smoothing}

\author{Cheng Chen \\
Department of ECE \\
University of Utah \\
\texttt{u0952128@utah.edu} \\
\And
Bhavya Kailkhura \\
Lawrence Livermore National Laboratory \\
Livermore, US \\
\texttt{kailkhura1@llnl.gov}
\And
Ryan Goldhahn \\
Lawrence Livermore National Laboratory \\
Livermore, US \\
\texttt{goldhahn1@llnl.gov}
\And
Yi Zhou \\
Department of ECE \\
University of Utah \\
\texttt{yi.zhou@utah.edu}
}

\iclrfinalcopy % Uncomment for camera-ready version, but NOT for submission.
\begin{document}

\maketitle

\begin{abstract}
Federated learning is an emerging data-private distributed learning framework, which, however, is vulnerable to adversarial attacks. Although several heuristic defenses are proposed to enhance the robustness of federated learning, they do not provide certifiable robustness guarantees. In this paper, we incorporate randomized smoothing techniques into federated adversarial training to enable data-private distributed learning with certifiable robustness to test-time adversarial perturbations. Our experiments show that such an advanced federated adversarial learning framework can deliver models as robust as those trained by the centralized training. Further, this enables provably-robust classifiers to $\ell_2$-bounded adversarial perturbations in a distributed setup. 
{We also show that one-point gradient estimation based training approach is {$2-3\times$ faster} than popular stochastic estimator based approach without any noticeable certified robustness differences.}
%{\color{red} (After discussion, two-point estimator has no advantage over one-point, so we don't present it. Contribution is applying adversarial smoothed training to federated learning and empirical exploration.)}
\end{abstract}

\vspace{-1mm}
\section{Introduction}\label{intro}
\vspace{-1mm}

Federated learning is an emerging distributed learning framework that enables edge computing at a large scale \citep{konevcny2016federated,li2020federated,mcmahan2016communication,chen2020fedcluster}, and has been successfully applied to various areas such as Internet of Things (IoT), autonomous driving, health care \citep{li2020federated}, etc. In particular, federated learning aims to exploit the distributed computation and heterogeneous data of a large number of edge devices to perform distributed learning while preserving full data privacy. The original federated learning framework proposed the federated averaging (FedAvg) algorithm \citep{mcmahan2016communication}. In each learning round, a subset of edge devices are selected to download a global model from the cloud server, based on which the selected devices train their local models using local data for multiple stochastic gradient descent (SGD) iterations. Then, these devices upload the trained local models to the server, where the local models are aggregated and averaged to obtain an updated global model that will be used in the next learning round. Throughout the federated learning process, all data are kept privately on the local devices. 
%However, as shown theoretically and empirically in [4, 5], the device-level data heterogeneity of the devices may slow down the convergence of FedAvg. \\

However, as modern federated learning often adopts over-parameterized models (e.g., deep neural networks) that have been proven to be vulnerable to adversarial perturbations to the test data \citep{Szegedy2014,Goodfellow2015,bulusu2020anomalous}, there is a rising concern about the adversarial robustness of the federated learning models used by massive number of edge devices. 
As an example, if a federated-trained model is vulnerable to adversarial examples, then its performance on edge devices solving safety-critical tasks can be significantly degraded in turn having serious consequences. 
%to classify an image $x$, a neural network model can be fooled by an adversary to misclassify $x+\delta$ where $\delta$ is an adversarial perturbation which is too small for human eyes to distinguish $x$ and $x+\delta$. 
To defend such adversarial attacks in federated learning, many studies propose to include standard adversarial training in the local training steps of federated learning \citep{zhou2021adversarially,zizzo2020fat,kerkouche2020federated,bhagoji19a}.  
%Some heuristic defenses are helpful for neural network models to resist such adversarial perturbations but insufficient for more powerful adversaries [8, 9, 10]. 
However, these approaches may not be able to defend strong adversaries and do not have certifiable adversarial robustness guarantee. To address these issues, some studies proposed the randomized smoothing technique that can train certifiably robust models at scale \citep{Lecuyer2019,cohen19c,li2020provable}. 

Specifically, randomized smoothing procedure uses a smoothed version of the original  classifier $f$ and certifies the adversarial robustness of the new classifier. The smoothed classifier is defined as $g(x) = \arg \max_c \mathbb{P}_{\delta \sim \mathcal{N}(0, \sigma^2I)}(f(x+\delta)=c)$, meaning the label of a data sample $x$ corresponds to the class whose decision region $\{x' \in \mathbb{R}^d: f(x') = c\}$ has the largest measure under the distribution $\mathcal{N}(x, \sigma^2I)$, where $\sigma$ is used for smoothing. 
Suppose that while classifying a point $\mathcal{N}(x, \sigma^2I)$, the original classifier $f$ returns the class $c_A$ with probability $p_A = \mathbb{P}(f(x + \delta) = c_A)$, and the “runner-up” class $c_B$ is returned with probability $p_B = \max_{c \neq c_A} \mathbb{P}(f(x + \delta) = c)$, then the prediction of the point $x$ under the smoothed classifier $g$ is robust within the radius $r(g;\sigma) = \frac{\sigma}{2}(\Phi^{-1}(p_A) - \Phi^{-1}(p_B)),$ where $\Phi^{-1}$ is the inverse CDF of the standard Normal distribution. In practice, Monte Carlo sampling is used to estimate a lower bound on $p_A$ and an upper bound on $p_B$ as its difficult to estimate the actual values for $p_A$ and $p_B$.
Since standard training of the base classifier does not achieve high robustness guarantees, \citep{cohen19c} proposed to use Gaussian data augmentation based training in which the base classifier is trained on Gaussian noise corruptions of the clean data.
%Specifically, randomized smoothing technique transforms a deep model $f$ into a smoothed model $g$ that has certifiable $l_2$ robustness guarantee. The smoothed model $g$ first perturbs a data sample $x$ using Gaussian noises and then use the original model $f$ to predict the labels for the perturbed samples. After that, a majority vote is performed to determine the final predicted label for this data sample $x$.
Recently, the authors in \citep{Salman2019} combined adversarial training approach with randomized smoothing to obtain significantly improved certification guarantees.

%To apply prior certifiable defenses to large neural networks, the randomized smoothing approach is introduced by  to transform a large neural network model $f$ into a new smoothed model $g$ that has certifiable $l_2$-norm robustness guarantees, which works as follows. Given an input $x$ in $\R^d$, the arbitrary base model $f$ maps it to a class. Meanwhile, the smoothed model $g(x)$ labels $x$ to class $c$ which is the most likely to be returned by the base model $f(x+\delta)$ with a noisy input $x+\delta$, where $\delta \sim \mathcal{N}(x, \sigma^2\mI)$. 
Such a smoothed model has been shown to outperform other existing certifiably robust models \citep{cohen19c} and the randomized smoothing scheme is applicable to deep networks and large datasets. To further enhance certifiable robustness of deep models, \citep{Salman2019} proposed an adversarial training approach that uses strong attacks generated against the smoothed model to train the smoothed model. In particular, \citep{Salman2019} demonstrated that such an adversarial training approach can substantially improve the robustness of smoothed models. However, these certifiably-robust training approaches are only applied to centralized learning setup, and similar provably-robust approaches in a federated learning setup is virtually non-existent. 
To bridge this gap, in this paper, we incorporate the randomized smoothing (with adversarial training) approach into the paradigm of federated learning to develop certifiably robust federated learning models. 
\textbf{Our contributions.} We apply the randomized smoothing (with adversarial training) approach to enable the certifiable robustness of federated learning to adversarial perturbations. Specifically, in the local training phase, each device applies adversarial training to train a robust smoothed local model to defend $\ell_2$ adversarial attacks. These local models are further aggregated by the central server to obtain a robust global model. To the best of our knowledge, this is the first work in the direction of enabling certifiable robust federated learning.

\vspace{-1mm}
\section{Federated Adversarial Learning with Randomized Smoothing}\label{algo}
\vspace{-1mm}

\subsection{Adversarial Learning with Randomized Smoothing}\label{smoothing}
\vspace{-1mm}

Consider a standard soft classifier $F_\theta$ that is parameterized by $\theta$ and maps an input data $x\in \mathbb{R}^d$ to a probability mass of class labels $\mathcal{Y}$. Then, its corresponding smoothed soft classifier $G_\theta$ is defined as 
\begin{align}
    G_\theta(x):= \mathbb{E}_{\delta \sim \mathcal{N}(0, \sigma^2I)}[F_\theta(x+\delta)].
\end{align}
Intuitively, the smoothed classifier $G_\theta$ perturbs the input sample with Gaussian noises and averages the predicted class distributions of all corrupted samples.
In particular, the standard deviation $\sigma$ of the Gaussian noise controls the level of certifiable robustness of the smoothed classifier.

To improve the performance, in \citep{Salman2019}, the authors proposed to leverage adversarial examples of the input data against the smoothed classifier $G_\theta$ (instead of $F_\theta$). Specifically, \citep{Salman2019} proposed the following adversarial training problem, where the training uses the adversarial data $\widehat{x}$ that is found within an $\ell_2$ ball of the original data $x$ by attacking $G_\theta$.
\begin{align} 
\textbf{SmoothAdv}:\quad \min_{\theta}\max_{\left\|\widehat{x}-x\right\|_{2} \leq \epsilon} J_\theta(\widehat{x}) :=-\log [G_\theta(\widehat{x})]_y, \label{eq: adv}
\end{align}
where $[G_\theta(\widehat{x})]_y$ denotes the $y$-th entry of the predicted classification probability mass. This approach is referred to as {\bf SmoothAdv} and the objective function is highly stochastic and non-convex. To solve the above adversarial optimization problem, two approaches were proposed in \citep{Salman2019}. For the first approach, the authors approximate the gradient of the above objective function using stochastic samples as follows 
\begin{align}
   \text{(Stochastic estimator)}\quad \nabla_{x} J (\widehat{x}) \approx -\nabla_{x}\log \bigg(\frac{1}{m} \sum_{i=1}^{m} [F_\theta \left(\widehat{x}+\delta_{i}\right)]_{y}\bigg),\label{eq: est1}
\end{align}
where $\delta_i, i=1,...,m$ are drawn i.i.d from $\mathcal{N}\left(0, \sigma^{2} I\right)$. 
Then, standard projected gradient ascent is applied to find adversarial samples. While the above stochastic gradient estimator provides an accurate gradient estimation, it is computational expensive as for every sample $x$ we need to perform back-propagation on a mini-batch of $m$ corrupted samples. 

To avoid performing back-propagation, \citep{Salman2019} discussed another gradient-free~\citep{liu2020primer} approach. Specifically, note that the adversarial optimization problem is equivalent to $\widehat{x} = {\arg \min }_{\left\|\widehat{x}-x\right\|_{2} \leq \epsilon} \big[G_\theta(\widehat{x})\big]_y$.
In particular, the gradient of $[G(\widehat{x})]_y$ can be conveniently characterized using the following one-point gradient-free estimator.
\begin{align} 
&\text{(One-point estimator)}\quad \nabla_{x}\big[G_\theta\left(\widehat{x}\right)\big]_{y} 
%= {\mathbb{E}}_{\delta \sim \mathcal{N}\left(0, \sigma^{2} I\right)} \left[\frac{\delta}{\sigma^{2}} \cdot [F_\theta\left(\widehat{x}+\delta\right)]_{y}\right]
\approx \frac{1}{m}\sum_{i=1}^m \left[\frac{\delta_i}{\sigma^{2}} \cdot [F_\theta\left(\widehat{x}+\delta_i\right)]_{y}\right]. \label{eq: est2}
\end{align}
The above estimator only involves function values that can be efficiently computed via forward-propagation. In particular, each gradient estimate $\frac{\delta_i}{\sigma^{2}} \cdot [F_\theta\left(\widehat{x}+\delta_i\right)]_{y}$ only needs to evaluate the function value at a single point $\widehat{x}+\delta_i$. 
Compared to the gradient-based stochastic estimator, this one-point estimator is computation lighter but induces a higher estimation variance. In \citep{Salman2019}, the performance of the one-point estimator was not evaluated for {\bf SmoothAdv}, and its comparison with the stochastic estimator was not comprehensive.

%\BK{If we only have one-point GE, we may want to change this sentence.}

% To overcome the issues of the previous two estimators, {\bf we propose} to use the following two-point gradient-free estimator.
% \begin{align} 
% &\text{(Two-point estimator)}\quad \nabla_{x^{\prime}}\big[G\left(x^{\prime}\right)_{y}\big] \approx \frac{1}{m}\sum_{i=1}^m \left[\frac{\delta_i}{\sigma^{2}} \Big[F\left(x^{\prime}+\delta_i\right)_{y} - F\left(x^{\prime}\right)_{y} \Big]\right]. \label{eq: est3}
% \end{align}
% It is easy to verify that the above estimator has the same mean as the one-point estimator. Moreover, it is based on the function value difference $F\left(x^{\prime}+\delta_i\right)_{y} - F\left(x^{\prime}\right)_{y}$, which can significantly reduce the estimation variance compared to the one-point estimator. 

\vspace{-1mm}
\subsection{Federated Adversarial Learning}
\label{fl_smoothadv}
\vspace{-1mm}

In this section, we incorporate the {\bf SmoothAdv} method into the federated learning framework. Our proposed algorithm is referred to as {\bf Fed-SmoothAdv} is presented in Algorithm \ref{algo: fedavg_smoothadv}.

\begin{algorithm}
%\vspace{-1mm}
	\caption{Federated Adversarial Learning ({\bf Fed-SmoothAdv})}\label{algo: fedavg_smoothadv}
	%\textbf{Hyperparameters:} $w_{0}$,  \\
    \textbf{Central-server executes:} \textit{$\quad \#$ Run on the central server} \\
    \For{\normalfont{learning round} $t=1,2, \ldots$}{
        Sample a subset $S_{t}$ of clients \\ 
        \For{\normalfont{each client} $k \in S_{t}$ \normalfont{in parallel}}{
            $\theta_{t+1}^{k} \leftarrow$ {\bf LocalTrain} $\left(k, \theta_{t}\right)$ \\
            Send $\theta_{t+1}^{k}$ to the server
            }
        Server aggregates $\theta_{t+1} \leftarrow \sum_{k\in S_t} \frac{n_{k}}{n} \theta_{t+1}^{k}$
        } 
\hrulefill\\
    \textbf{LocalTrain} $(k, \theta):$ \textit{$\quad \#$ Local training of client $k$} \\
        \For{\normalfont{local iteration} $i=1,2,...,E$}{
            Sample a minibatch of data $b$\\
            $\theta \leftarrow$ {\bf SmoothAdv} $\left(\theta, b\right)$ \textit{$\quad \#$ Use one of the two gradient estimators} 
            }
\hrulefill\\ 
    \textbf{SmoothAdv} $(\theta, b):$ \textit{$\quad \#$ Adversarial training with randomized smoothing} \\
    Data samples $\left(x^{(1)}, y^{(1)}\right),\left(x^{(2)}, y^{(2)}\right), \ldots,\left(x^{(b)}, y^{(b)}\right)$ \\
        
    Generate noises $\{\delta_{i}^{(j)}\}_{i=1}^m \sim \mathcal{N}\left(0, \sigma^{2} I\right)$ for any $x^{(j)}, j=1,...,b$ \\
        $L \leftarrow []$ \textit{$\quad \#$ List of adversarial examples} \\
        \For{$1 \leq j \leq b$}{
            Generate adversarial sample $\widehat{x}^{(j)}$ for ${x}^{(j)}$ by attacking the smoothed classifier using one of the gradient estimators in eqs.(\ref{eq: est1},\ref{eq: est2}) and noises $\{\delta_{i}^{(j)}\}_{i=1}^m$.\\

            Append $\{ (\widehat{x}^{(j)}+\delta_{1}^{(j)}, y^{(j)}), \ldots,(\widehat{x}^{(j)}+\delta_{m}^{(j)}, y^{(j)})\}$ to list $L$.
            }
    Train model $\theta$ using adversarial samples in $L$ for multiple SGD steps.
    \vspace{-1mm}
\end{algorithm}

To elaborate, the hierarchical structure of {\bf Fed-SmoothAdv} is the same as that of standard federated learning, i.e., a subset of edge devices is sampled in every round to perform local training, and then their local models are aggregated by the cloud server. 
However, in our federated adversarial learning, each client uses {\bf SmoothAdv} to perform local adversarial training using strong adversarial samples generated by attacking the smoothed local model.

\vspace{-3mm}
\section{Experiments}\label{exp}
\vspace{-2mm}

We compare the certified robustness of {\bf Fed-SmoothAdv} with the baseline method {\bf SmoothAdv} in training an AlexNet \citep{Krizhevsky2012} on CIFAR-10 \citep{Krizhevsky09}. Here, certified robustness is defined as the fraction of the test samples that are correctly classified (without abstaining) by $G_{\theta}$ and are certified within an $\ell_2$ radius of $r$. We set the smoothing parameter $\sigma=\{0.12, 0.25, 0.5\}$ and the perturbation bound $\epsilon=\{64, 128, 256\}$, and use the same $\sigma$ for certification as that used in the training.
For both methods, we apply both the stochastic estimator and the one-point estimator. Moreover, we test {\bf Fed-SmoothAdv} under different levels of device data heterogeneity $\gamma_{\text{device}}$ (the higher the more heterogeneous). Please refer to \Cref{append: hyper} for all the other hyperparameters used in the experiments.

%We report the upper envelope of the obtained certified accuracy by choosing the best model for each radius over a grid of hyperparameters (see Appendix~\ref{}). 

\begin{figure}[bth]
	\centering
	\vspace{-4mm}
	\includegraphics[width=0.32\textwidth,height=0.25\textwidth]{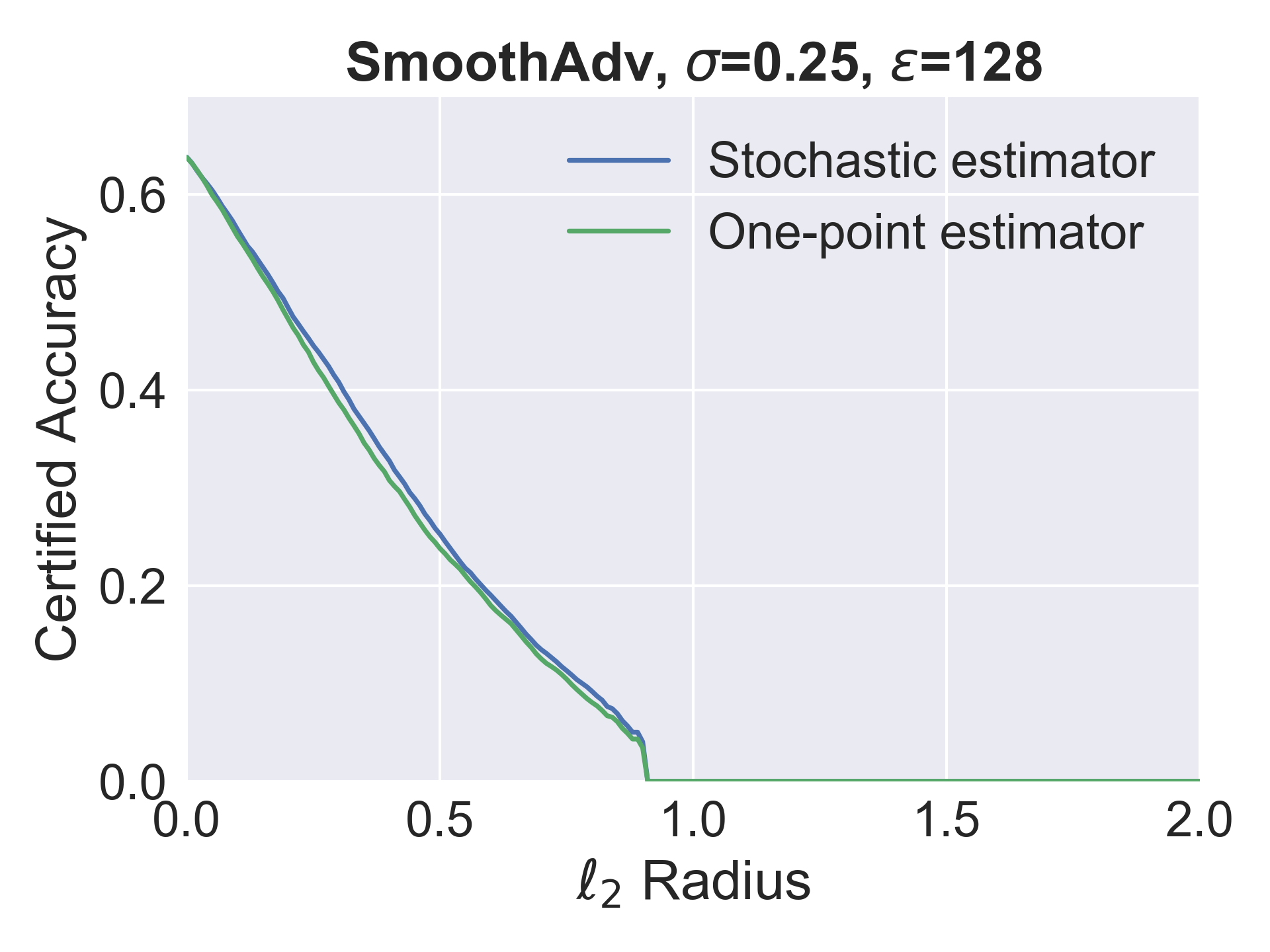}	
	\includegraphics[width=0.32\textwidth,height=0.25\textwidth]{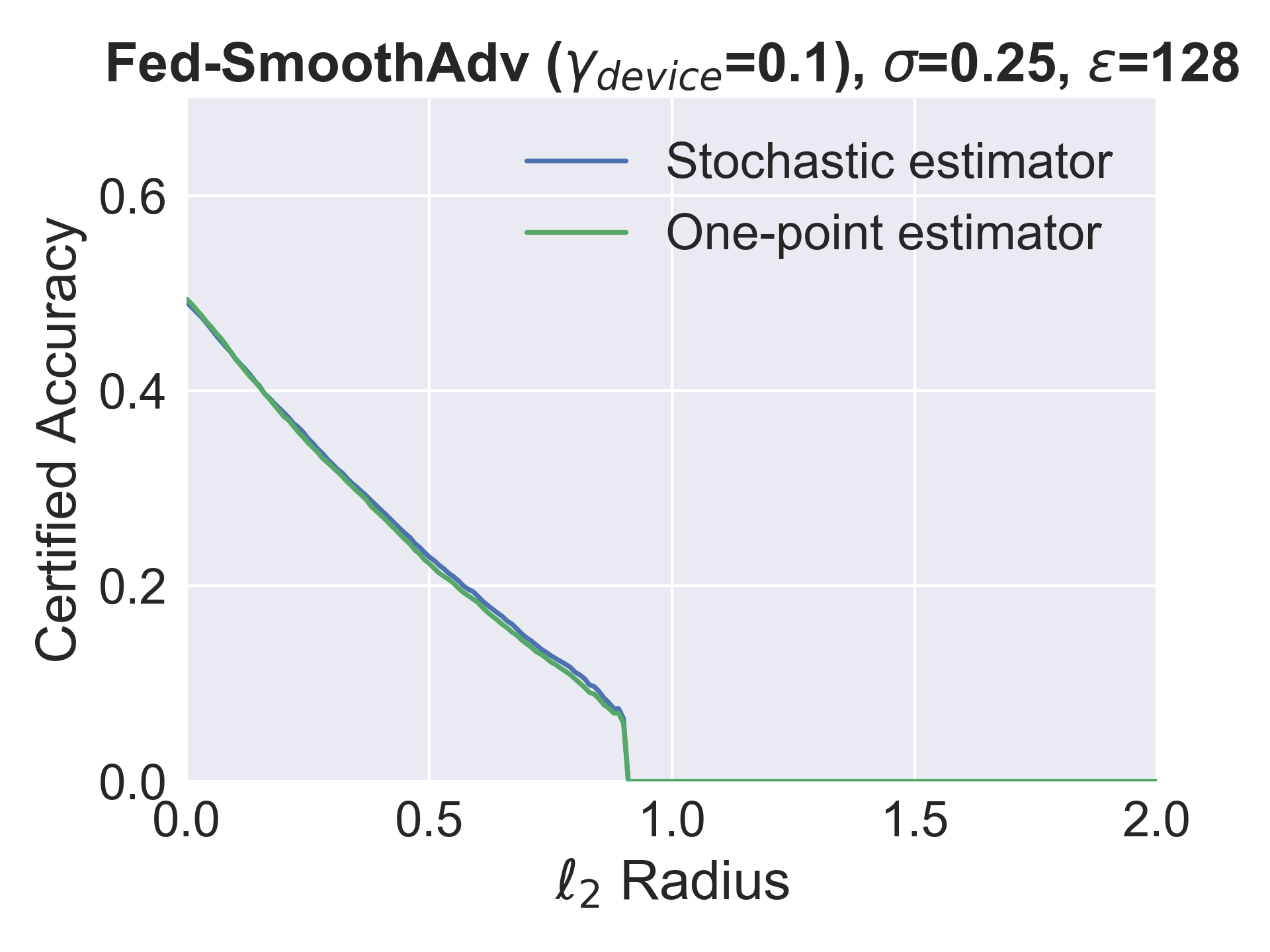}	
	\includegraphics[width=0.32\textwidth,height=0.25\textwidth]{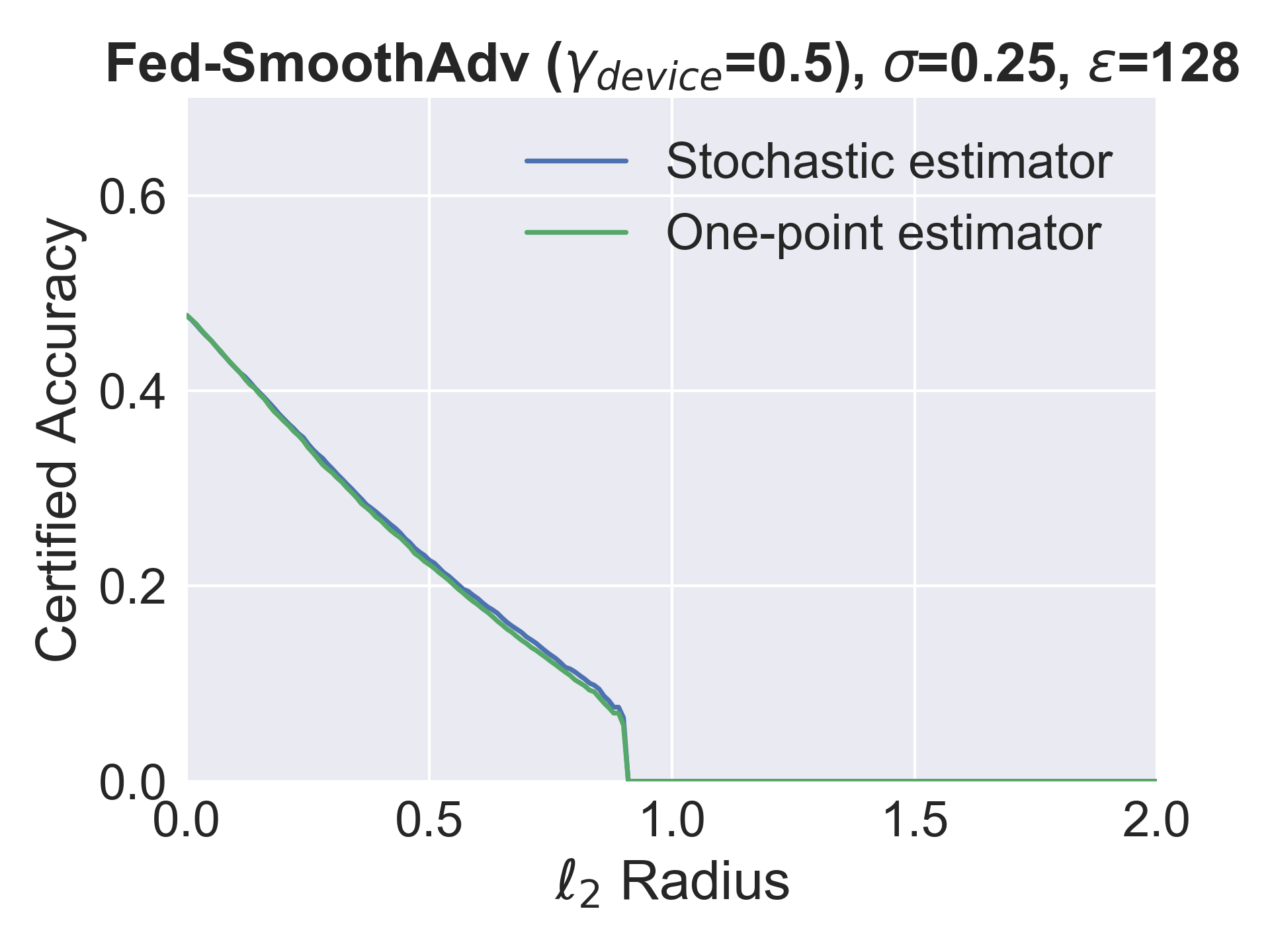}
	\vspace{-4mm}
	\caption{\small{Certified accuracy of {\bf SmoothAdv} and {\bf Fed-SmoothAdv} with $\sigma=0.25$ and $\epsilon=128$.}}\label{fig: 1}
	\vspace{-4mm}
\end{figure}

\begin{figure}[bth]
	\vspace{-2mm}
	\centering
	\includegraphics[width=0.32\textwidth,height=0.25\textwidth]{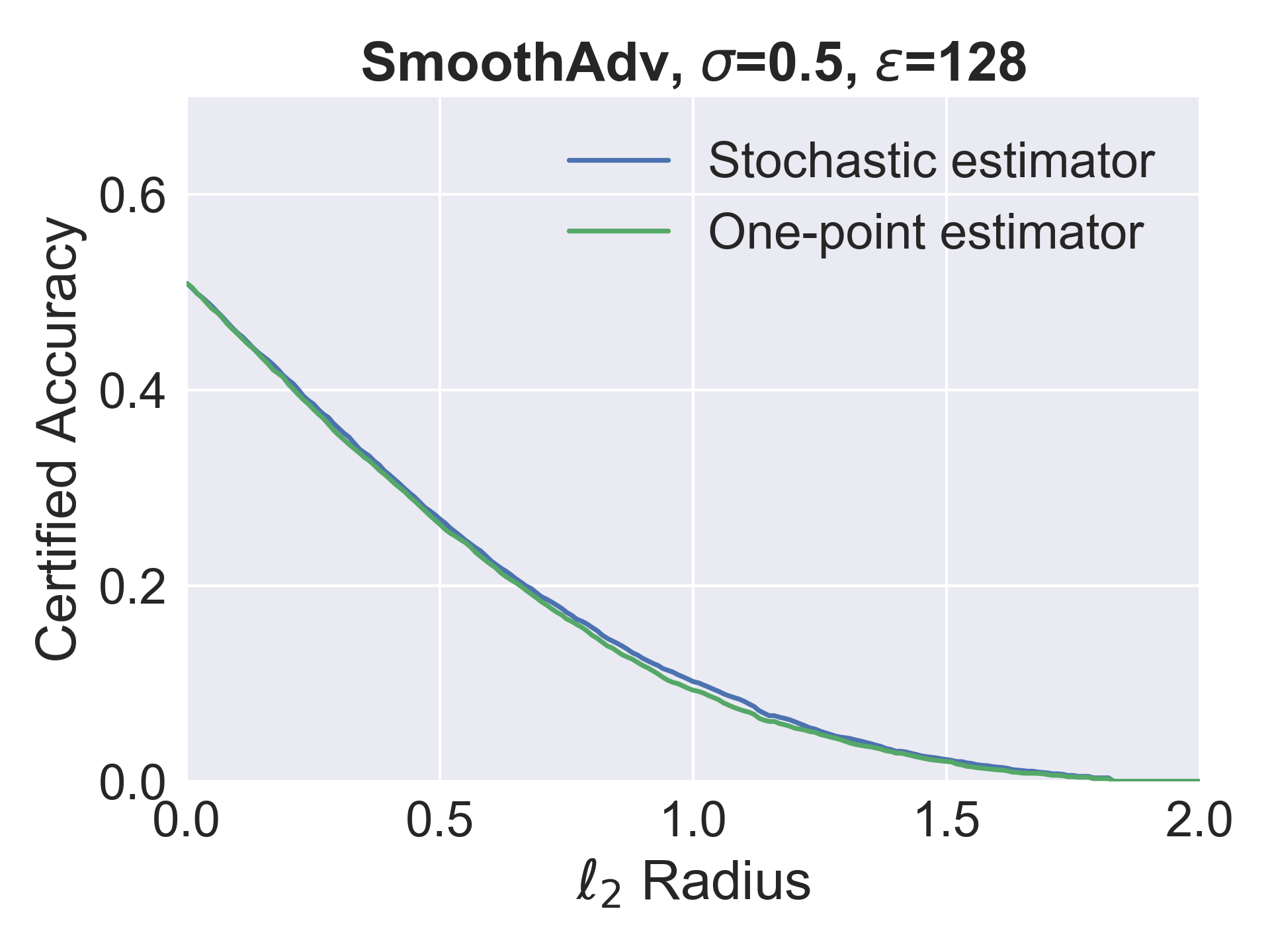}	
	\includegraphics[width=0.32\textwidth,height=0.25\textwidth]{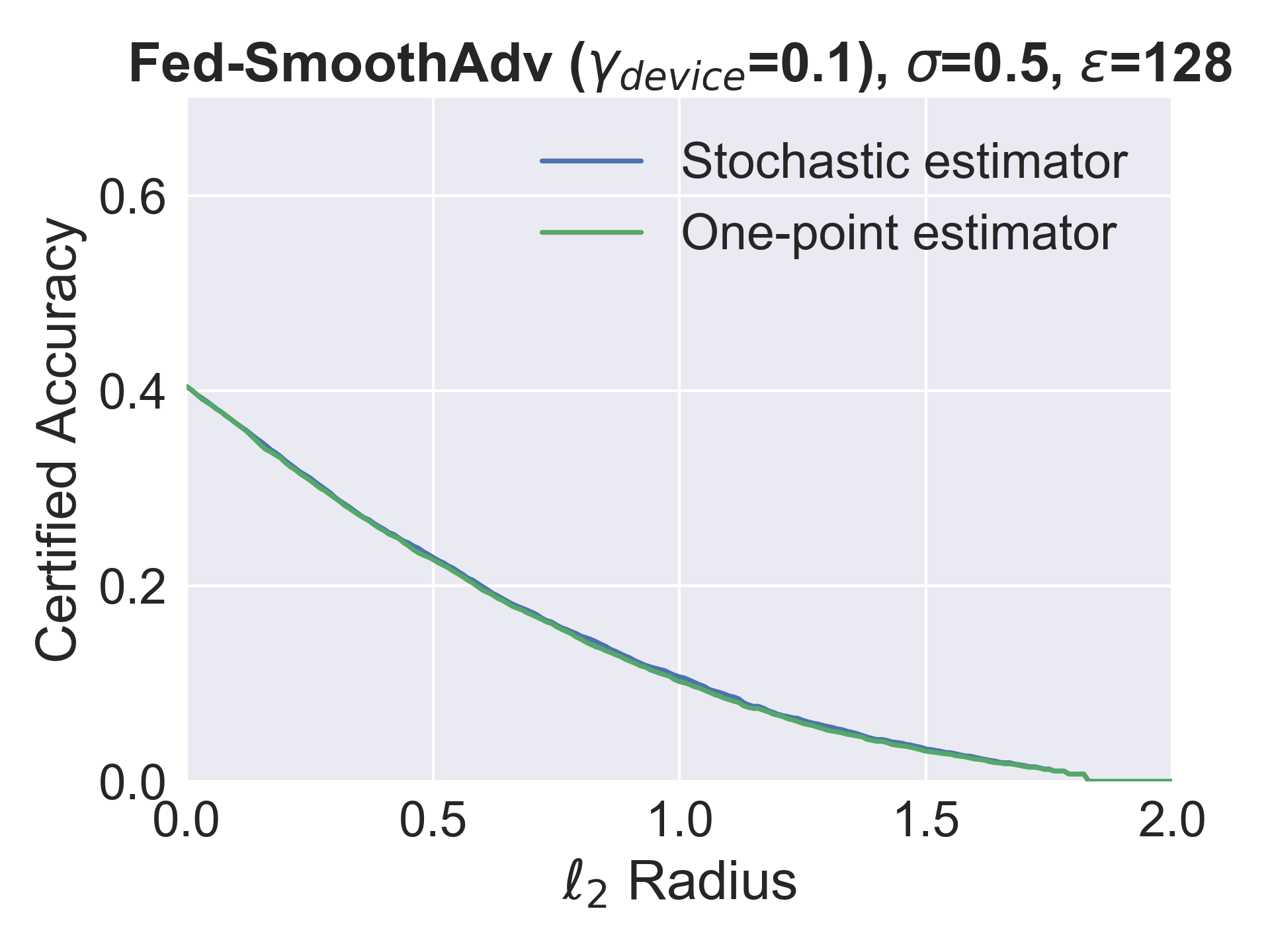}	
	\includegraphics[width=0.32\textwidth,height=0.25\textwidth]{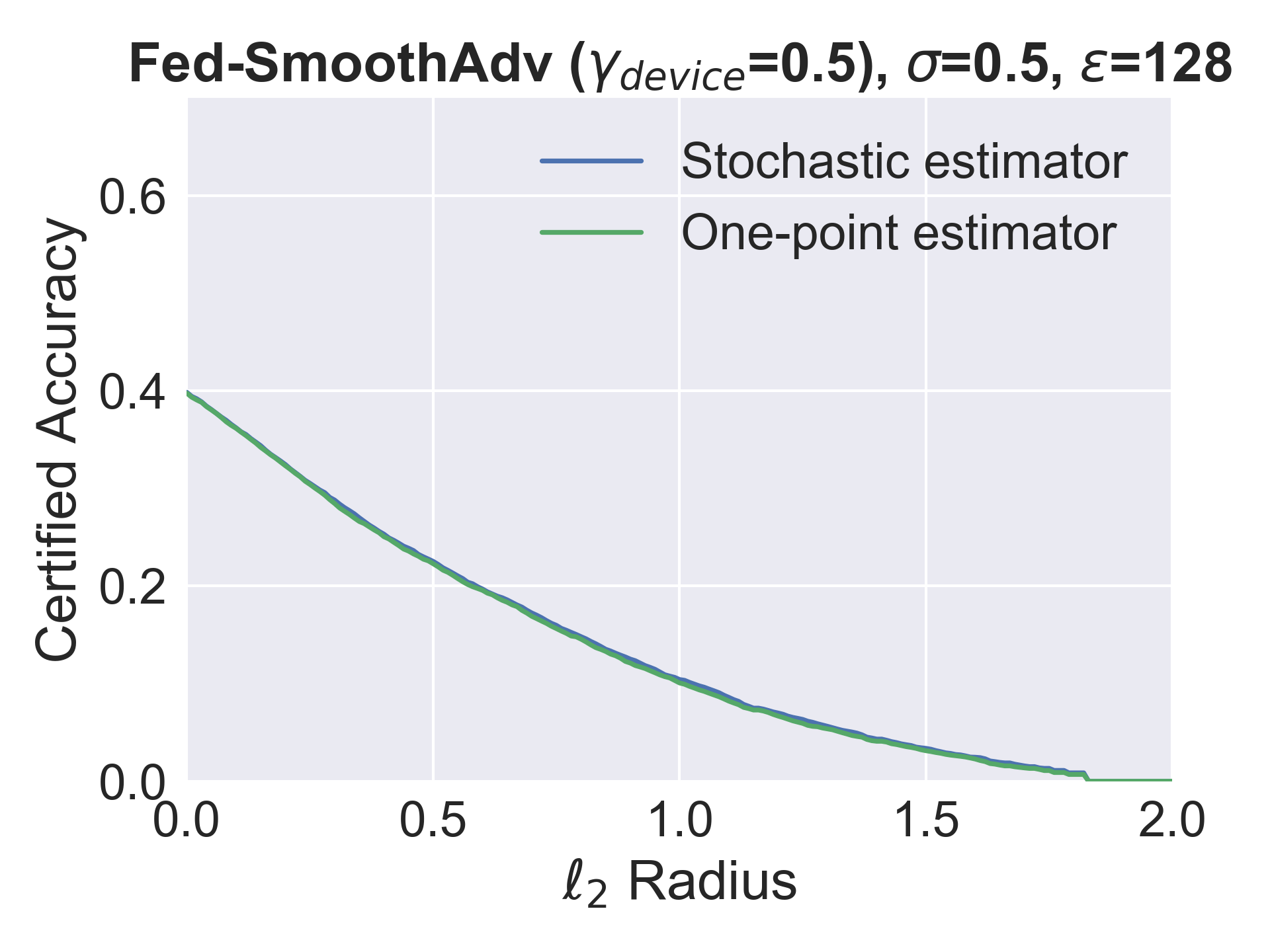}
	\vspace{-4mm}
	\caption{\small{Certified accuracy of {\bf SmoothAdv} and {\bf Fed-SmoothAdv} with $\sigma=0.5$ and $\epsilon=128$.}}\label{fig: 2}
	\vspace{-2mm}
\end{figure}

In \Cref{fig: 1}, we plot the certified accuracy of both {\bf SmoothAdv} and {\bf Fed-SmoothAdv} (with heterogeneity $\gamma_{\text{device}}=0.1, 0.5$) with $\sigma=0.25, \epsilon=128$. It can be seen that the certified accuracy of {\bf Fed-SmoothAdv} is slightly lower than that of {\bf SmoothAdv}, but is reasonably close. Also, the data heterogeneity $\gamma_{\text{device}}$ does not affect the certified accuracy of {\bf Fed-SmoothAdv}, which implies that {\bf SmoothAdv} can be effectively applied to enhance the adversarial robustness of heterogeneous federated learning. Moreover, we note that while the performance of the one-point estimator is almost the same as that of the stochastic estimator, the training time is significantly reduced by {2-3 times} due to avoidance of backpropagation. All these results show that applying {\bf SmoothAdv} with the one-point estimator to federated learning can efficiently enhance the certified model accuracy. In \Cref{fig: 2}, we plot the certified accuracy results under $\sigma=0.5$ and $\epsilon=128$. One can observe a similar comparison between the two methods as that in \Cref{fig: 1}. In particular, with a larger $\sigma$, the certified accuracy is lower but spans over a wider range of $\ell_2$ radius. {The results corresponding to other choices of $\sigma,\epsilon$ can be found in \Cref{append: add}.}

%From Figure \ref{fig: 2}, we observe the same trend as the results of subsection \ref{sigma0.25}. Therefore, the conclusions of subsection \ref{sigma0.25} are also applied here, which means $\sigma$ will not affect the main conclusions.

%However, there are also some differences between results of subsection \ref{sigma0.25} and \ref{sigma0.5}. By comparing results of the two subsections, we can observe that $\sigma$ controls a robustness/accuracy trade-off, as mentioned by [15]. Specifically, when $\sigma$ is low (i.e., $0.25$), small radii ($\leq 0.9$) can be certified with high accuracy, but large radii ($> 0.9$) cannot be certified at all. However, when $\sigma$ is high (i.e., $0.5$), larger radii ($> 0.9$) can be certified, but smaller radii are certified at a lower accuracy.

\begin{wrapfigure}[9]{R}{0.38\textwidth}
	\centering
	\vspace{-6mm}
	\includegraphics[width=0.32\textwidth,height=0.25\textwidth]{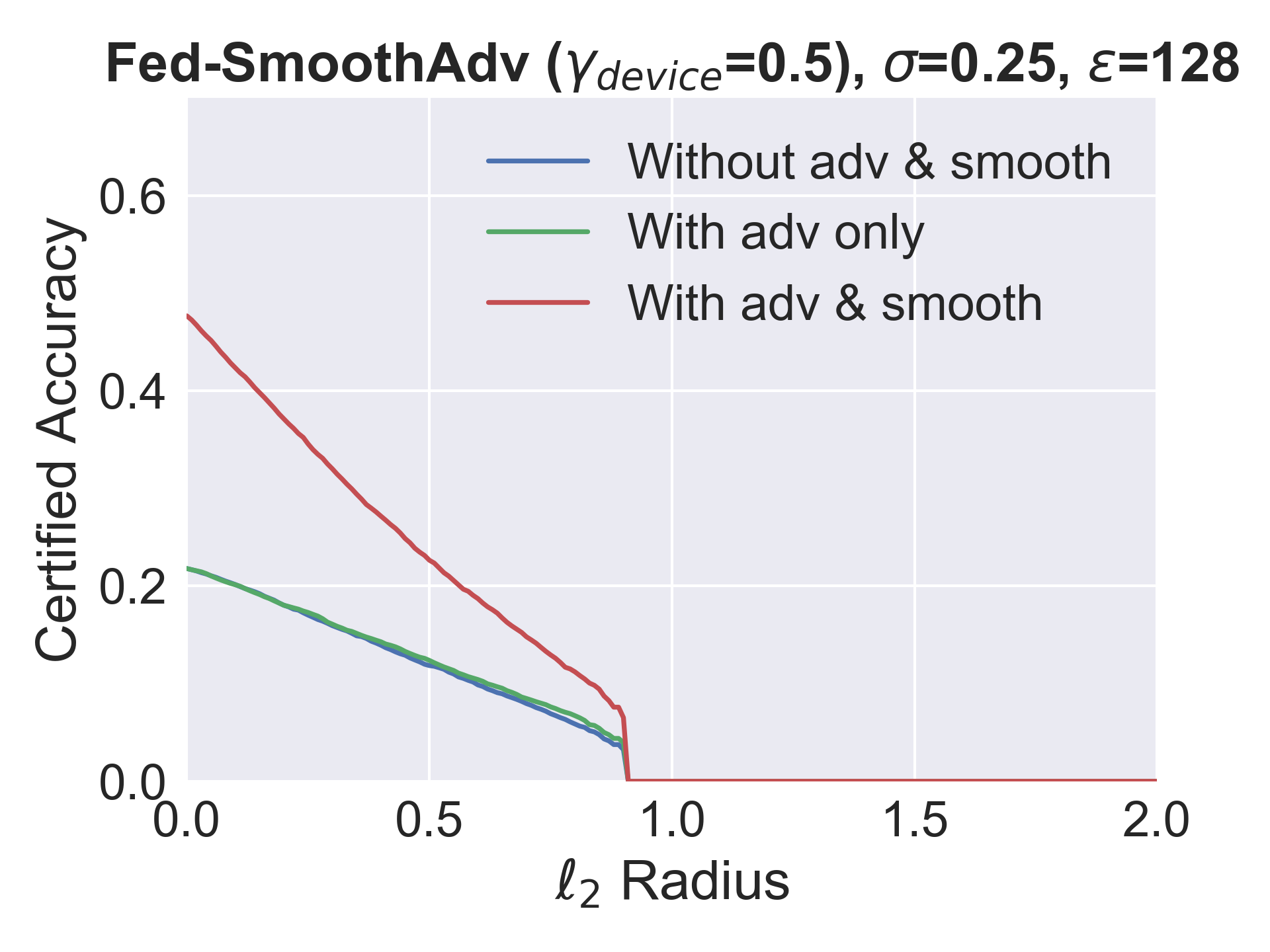}
		\vspace{-6mm}
	\caption{Ablation study of \small{{\bf Fed-SmoothAdv}.}}\label{fig: b11}
\end{wrapfigure}
We further explore the certified accuracy of {\bf Fed-SmoothAdv} under ablation settings.
\Cref{fig: b11} plots the result under $\sigma=0.25, \epsilon=128$ with heterogeneous data. Please see \Cref{append: base} for more details and results. 
It can be seen that adversarial training of smoothed classifiers is critical for achieving a high certified accuracy. Standard training and adversarial training of original classifier (without smoothing) perform poorly in terms of certified robustness. This demonstrates the necessity of smoothed classifier to enable certifiably-robust federated learning.

\vspace{-2mm}
\section{Conclusion}\label{conclusion}
\vspace{-2mm}

In this paper, we incorporated the randomized smoothing techniques into the federated adversarial learning framework to enable certifiable robustness to test-time adversarial perturbations. 
We demonstrated through extensive experiments that our adversarially smooth federated learning models could successfully achieve similar certified robustness as the centralized models. 
Meanwhile, we empirically proved that the device data heterogeneity and type of gradient estimator did not affect the performance much. 
%In addition, we also empirically certified the existence of robustness/accuracy trade-off in federated learning which is controlled by $\sigma$. 
The attempt in this paper is crucial for the applications of federated learning because of the adversarial attacks on its user's devices and the resulting strong demand for user's data privacy and security in the real world. 
In the future, we will apply randomized smoothing to more complex federated learning frameworks~\citep{chen2020fedcluster} and theoretically study its performance.

\section*{Acknowledgements} 
This work was performed under the auspices of the U.S. Department of Energy by the Lawrence Livermore National Laboratory under Contract No. DE-AC52-07NA27344, Lawrence Livermore National Security, LLC. This document was prepared as an account of the work sponsored by an agency of the United States Government. Neither the United States Government nor Lawrence Livermore National Security, LLC, nor any of their employees makes any warranty, expressed or implied, or assumes any legal liability or responsibility for the accuracy, completeness, or usefulness of any information, apparatus, product, or process disclosed, or represents that its use would not infringe privately owned rights. Reference herein to any specific commercial product, process, or service by trade name, trademark, manufacturer, or otherwise does not necessarily constitute or imply its endorsement, recommendation, or favoring by the United States Government or Lawrence Livermore National Security, LLC. The views and opinions of the authors expressed herein do not necessarily state or reflect those of the United States Government or Lawrence Livermore National Security, LLC, and shall not be used for advertising or product endorsement purposes. This work was supported by LLNL Laboratory Directed Research and Development project 20-SI-005 and released with LLNL tracking number LLNL-CONF-820514.

\bibliography{iclr2021_conference,ref_FedCycs}
\bibliographystyle{iclr2021_conference}

\newpage
\section*{Supplementary material}
\appendix
\section{Experiment Setup and hyperparameters}\label{append: hyper}

For {\bf Fed-SmoothAdv}, we simulate $1000$ edge devices and only $10\%$ of them are sampled in each learning round. Each device holds 500 data samples. To control the data heterogeneity of each device, we define a data heterogeneity ratio $\gamma_{\text{device}}\ \text{in } (0, 1)$. Specifically, we randomly assign one class label as the major class of each device. Then, for each device, $\gamma_{\text{device}}$ portion of samples are sampled from the major class, and the rest $(1-\gamma_{\text{device}})$ portion of samples are drawn from the remaining classes uniformly at random. In the experiments , we set $\gamma_{\text{device}}=0.1,0.5$ that correspond to homogeneous data and heterogeneous data, respectively.

In the experiments, we set the number of Gaussian noise samples to be $m=2$, and use $2$ projected gradient descent steps for generating the adversarial samples. We set the inner-learning-rate for generating adversarial samples to $0.01$, and the outer-learning-rate for updating the model parameters to $0.01$. We set batch-size to $30$ for each activated device of {\bf Fed-SmoothAdv} and $60$ for {\bf SmoothAdv}. Moreover, each activated device of {\bf Fed-SmoothAdv} uses $20$ batches of data in the local training of a learning round, and this is equivalent to $1000$ batches of data used by the centralized {\bf SmoothAdv}. The total number of learning rounds is $150$. In the certification phase, we set $\alpha=0.001$, which means that there is at most $0.1\%$ chance that the certification falsely certifies a non-robust input. 

\section{Ablation Studies}\label{append: base}
In this section, we explore the certified accuracy of {\bf Fed-SmoothAdv} 
%(with heterogeneity $\gamma_{\text{device}}=0.1, 0.5$) with $\sigma=0.25, 0.5$ and $\epsilon=128$ 
under the following ablation settings: (1) Without adv \& smooth: standard training of original classifier; (2) With adv only: adversarial training of original classifier (stochastic estimator); and (3) With adv \& smooth: adversarial training of smoothed classifier, i.e., {\bf{Smoothadv}} (stochastic estimator). 
%All the trained models in a given figure use the same $\sigma$ value to be certified so that we can evaluate their accuracy within the same radius.

\begin{figure}[bth]
	\vspace{-2mm}
	\centering
	\includegraphics[width=0.32\textwidth,height=0.25\textwidth]{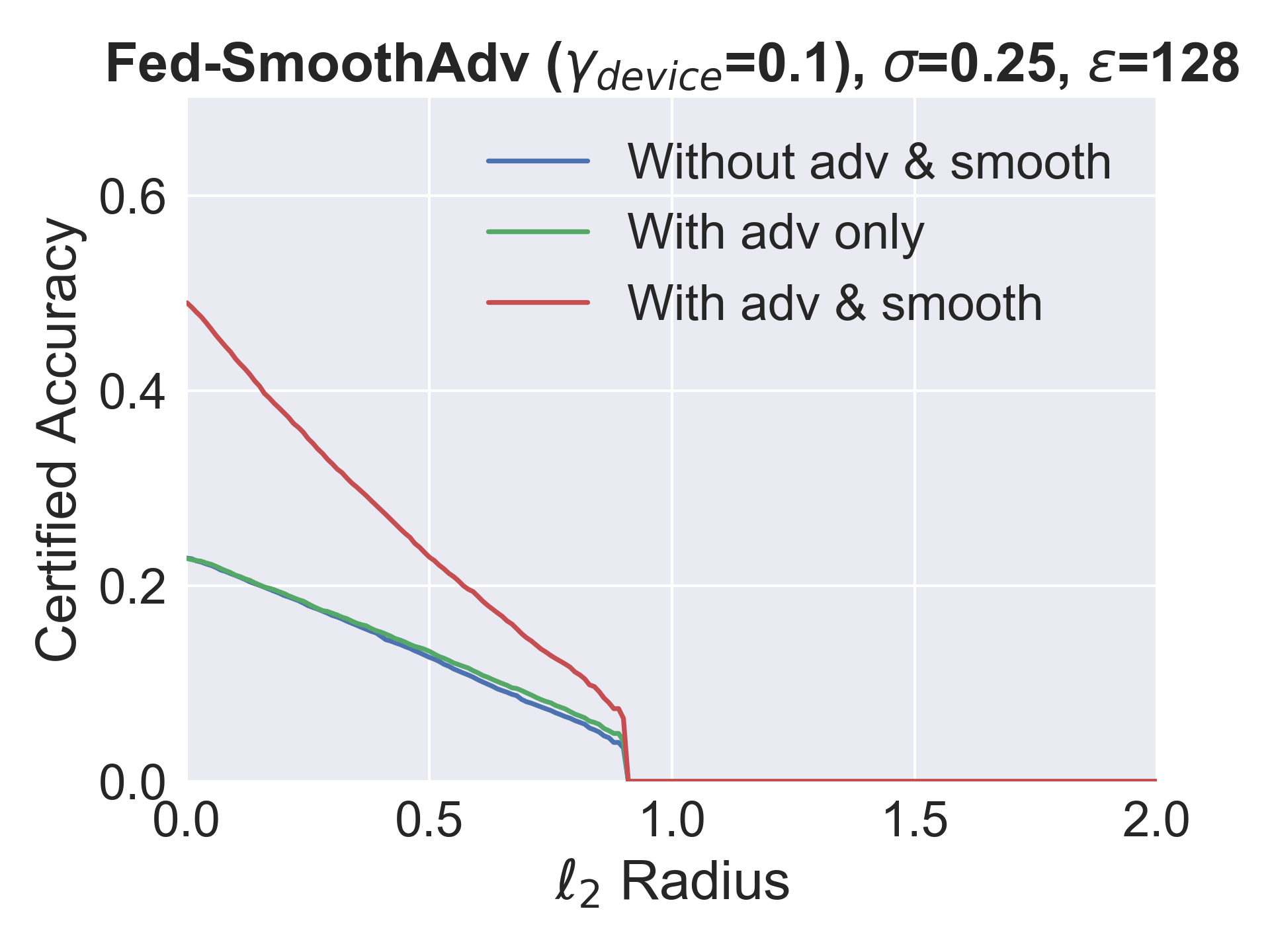}	
	\includegraphics[width=0.32\textwidth,height=0.25\textwidth]{Images/Appendix/fedavg5_sigma25_base.png}	
	\vspace{-4mm}
	\caption{\small{Ablation study of {\bf Fed-SmoothAdv} with $\sigma=0.25$ and $\epsilon=128$.}}\label{fig: b1}
	\vspace{-2mm}
\end{figure}

\begin{figure}[bth]
	\vspace{-2mm}
	\centering
	\includegraphics[width=0.32\textwidth,height=0.25\textwidth]{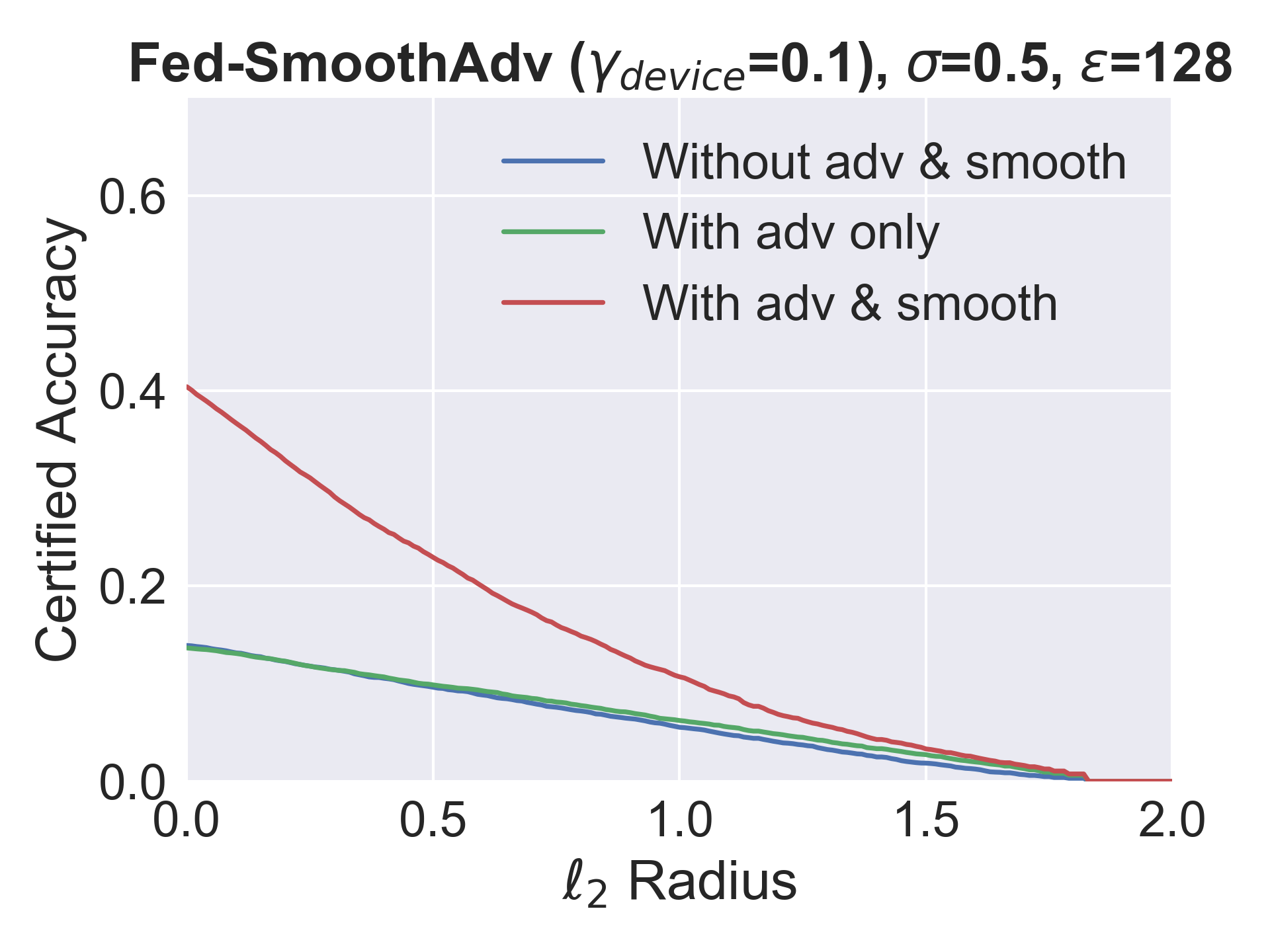}	
	\includegraphics[width=0.32\textwidth,height=0.25\textwidth]{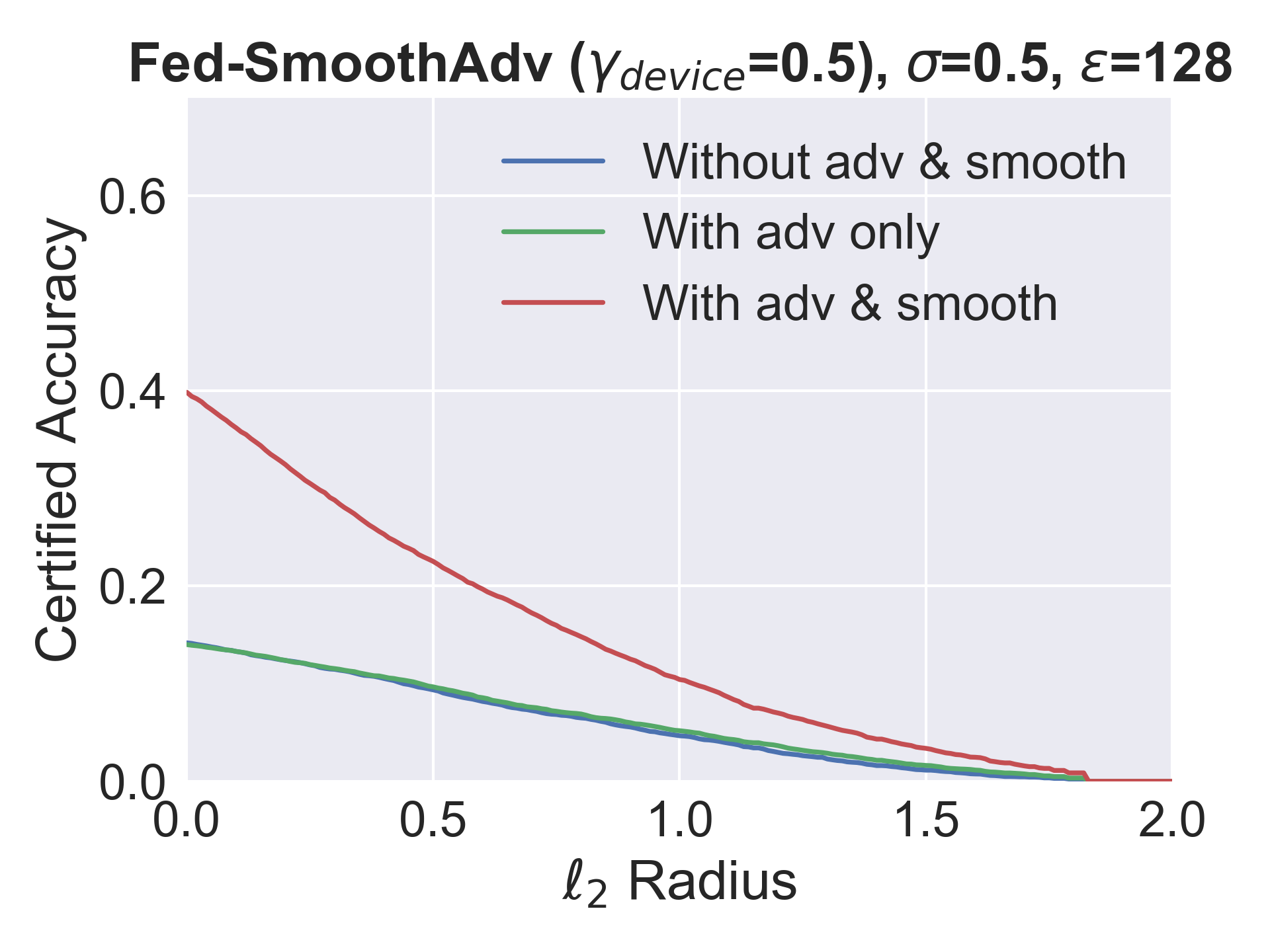}	
	\vspace{-4mm}
	\caption{\small{Ablation study of {\bf Fed-SmoothAdv} with $\sigma=0.5$ and $\epsilon=128$.}}\label{fig: b2}
	\vspace{-2mm}
\end{figure}

%The results are presented in \Cref{fig: b1} and \Cref{fig: b2}, where ``adv" refer to ``adversarial training". 
All of the plots yield following conclusions. First, the certified accuracy of {\bf Fed-SmoothAdv} is much higher than that of standard and adversarial training of original (non-smoothed) classifier, which indicates that randomized smoothing is very helpful to improve the performance of {\bf Fed-SmoothAdv}. Second, adversarial training does not achieve significantly higher certified accuracy than standard training, which again indicates that importance of having a smoothed classifier. 
%In summary, adversarial training is a little helpful to improve {\bf Fed-SmoothAdv}'s certified accuracy, while randomized smoothing is much more helpful to do so. This demonstrates the necessity to add randomized smoothing to certifiably-robust federated adversarial learning, which is what we proposed in this paper.

\section{Additional Experimental Results}\label{append: add}
\begin{figure}[bth]
	\vspace{2mm}
	\centering
	\includegraphics[width=0.32\textwidth,height=0.25\textwidth]{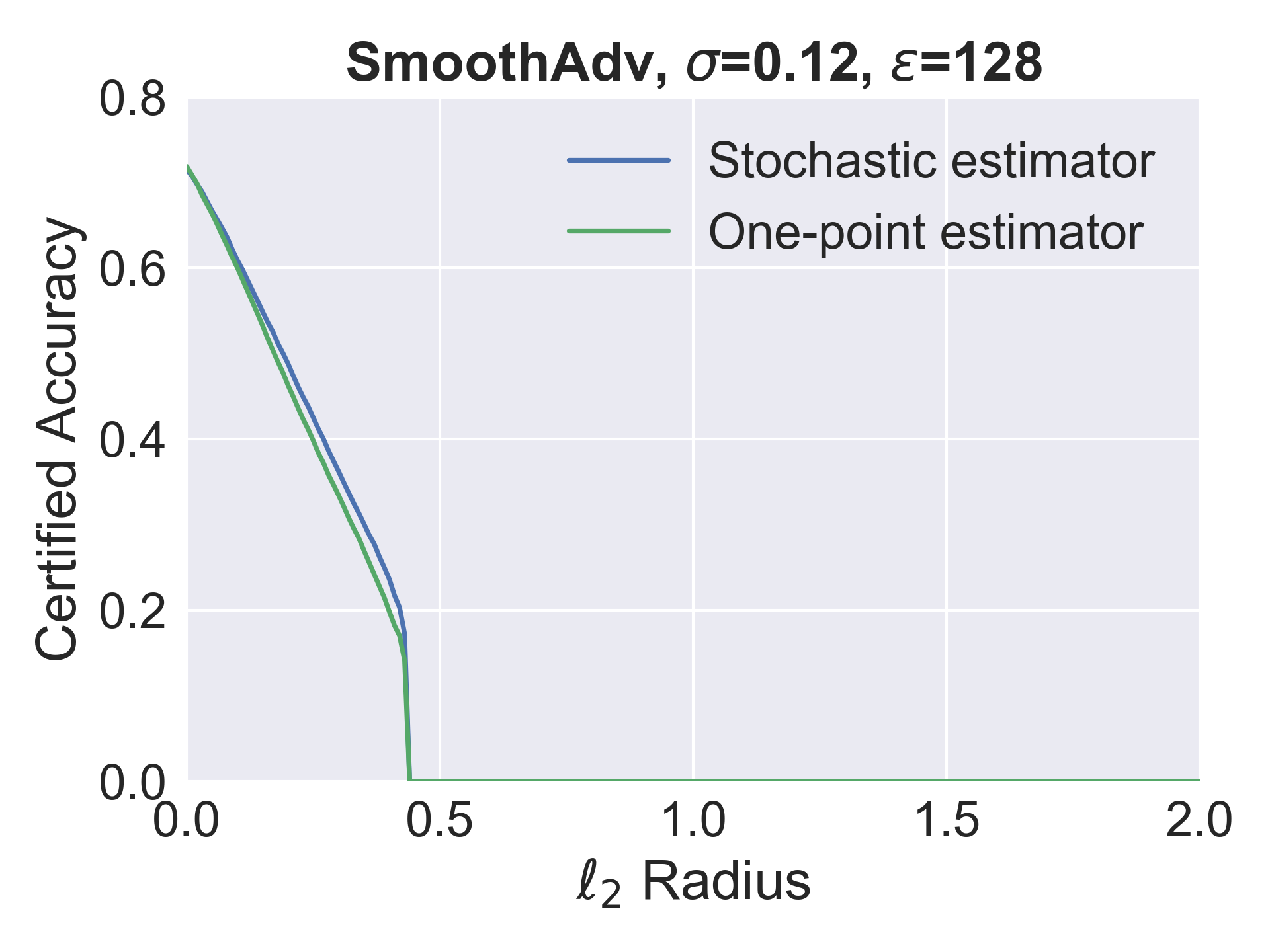}	
	\includegraphics[width=0.32\textwidth,height=0.25\textwidth]{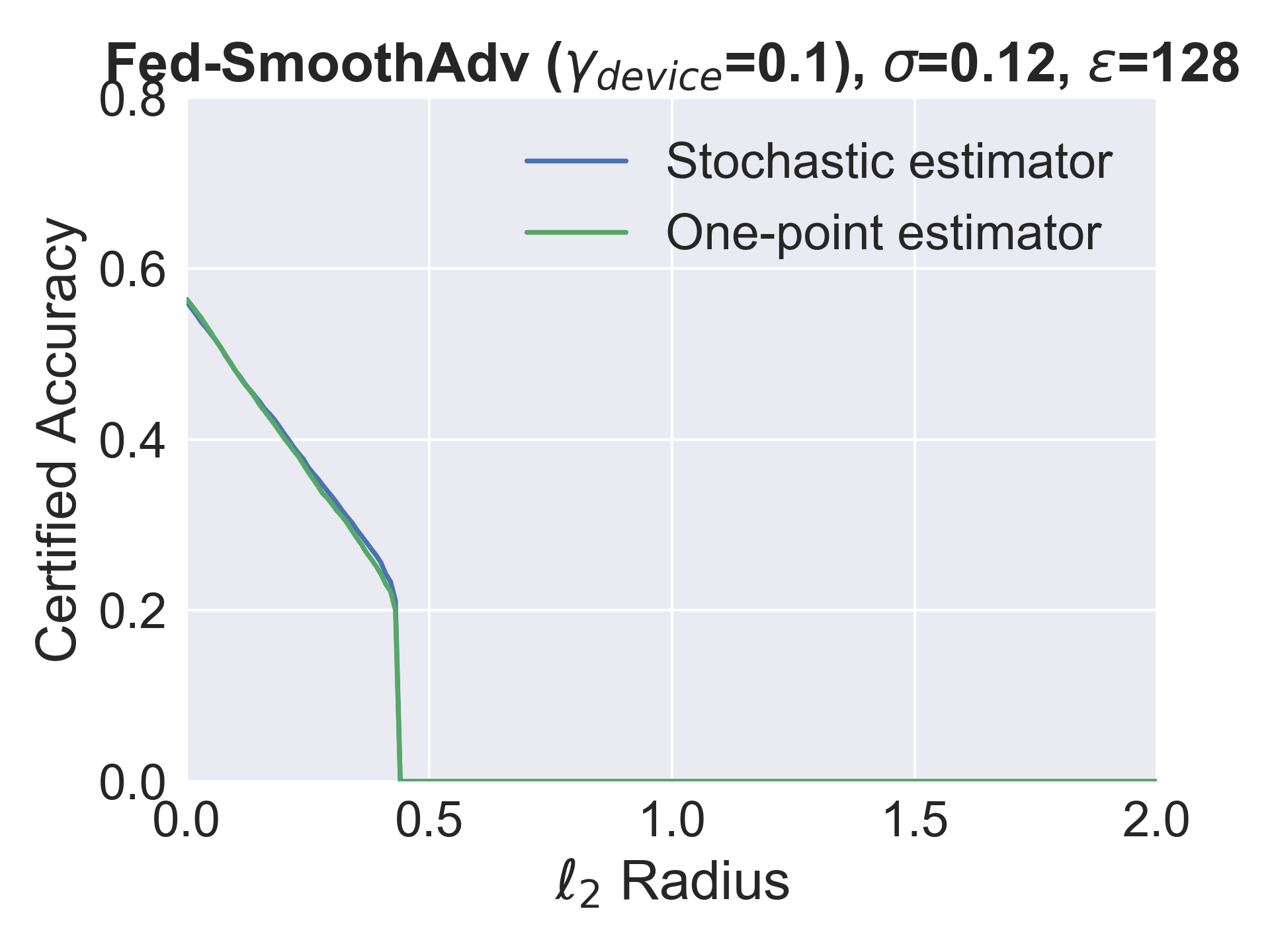}	
	\includegraphics[width=0.32\textwidth,height=0.25\textwidth]{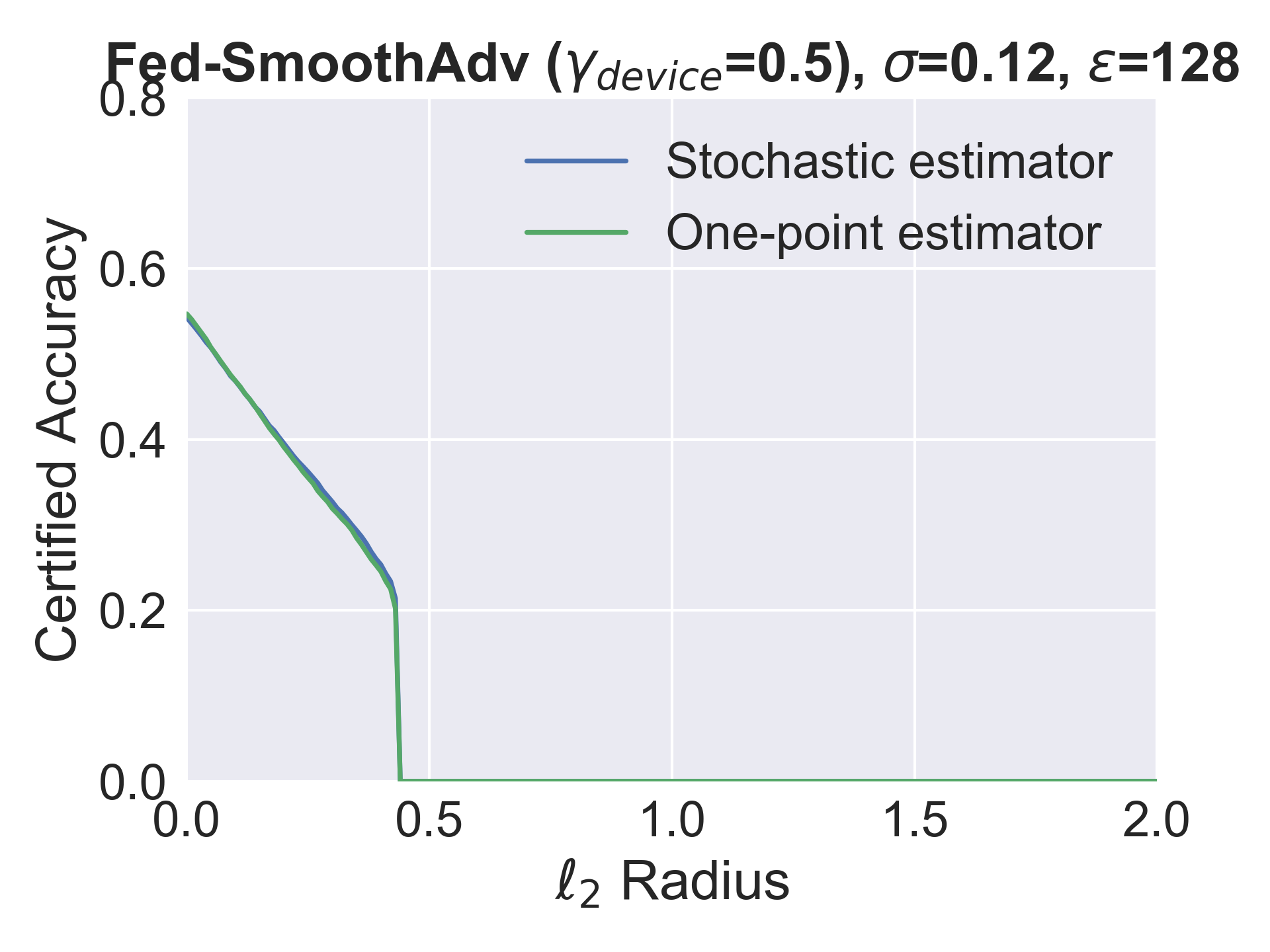}
	\vspace{-4mm}
	\caption{\small{Certified accuracy of {\bf SmoothAdv} and {\bf Fed-SmoothAdv} with $\sigma=0.12$ and $\epsilon=128$.}}\label{fig: c1}
	\vspace{-2mm}
\end{figure}

\begin{figure}[bth]
	\vspace{2mm}
	\centering
	\includegraphics[width=0.32\textwidth,height=0.25\textwidth]{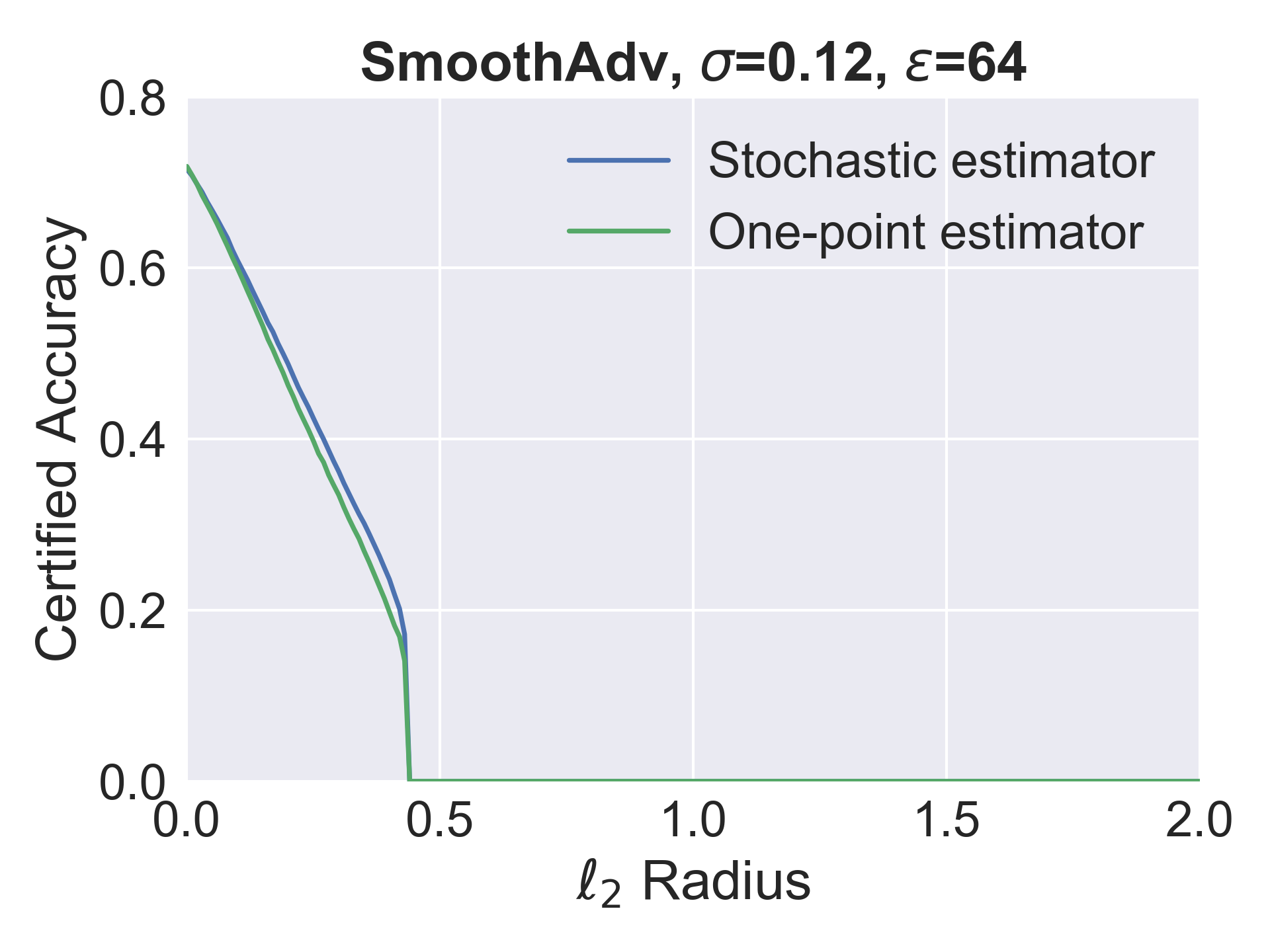}	
	\includegraphics[width=0.32\textwidth,height=0.25\textwidth]{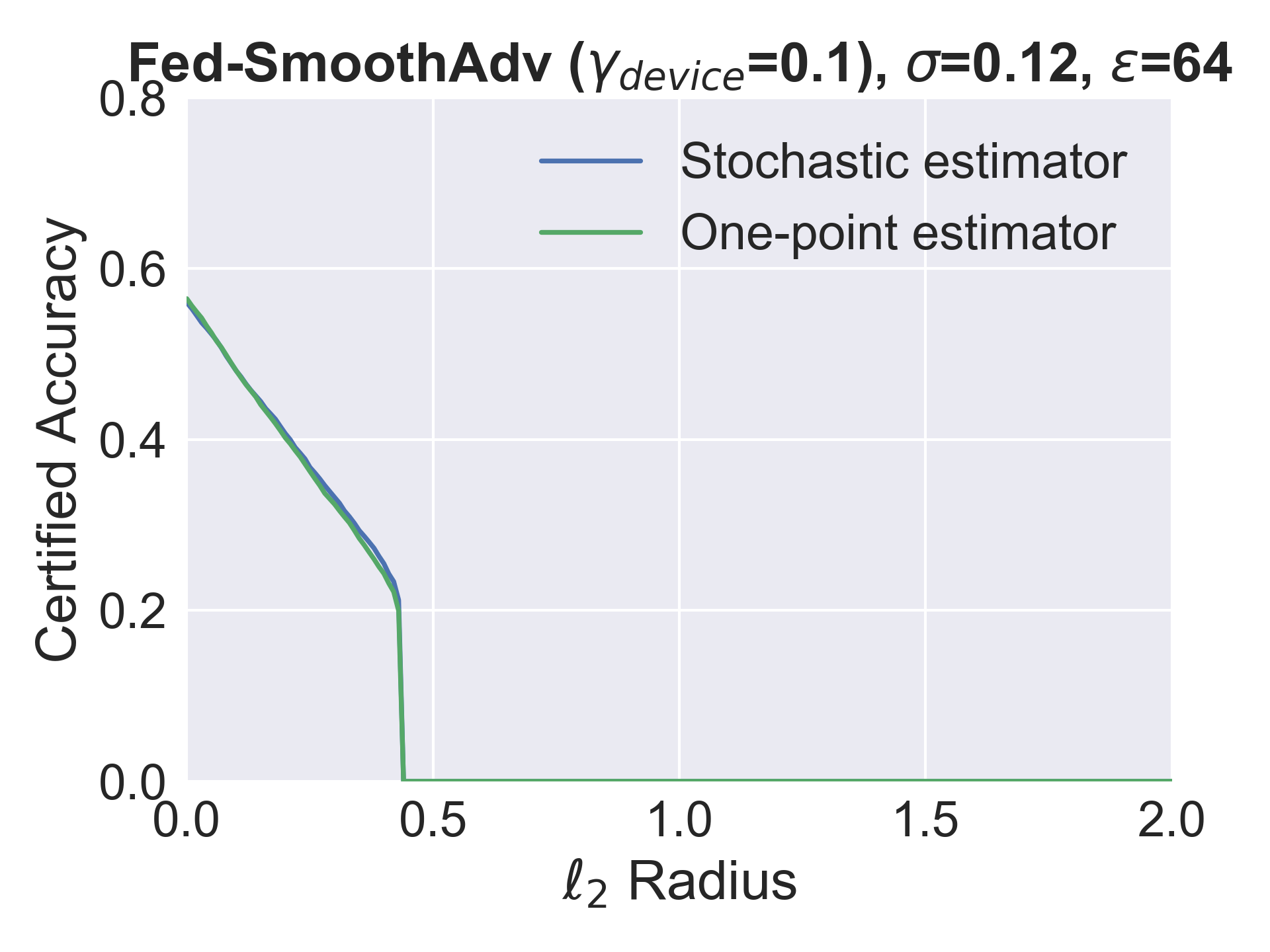}	
	\includegraphics[width=0.32\textwidth,height=0.25\textwidth]{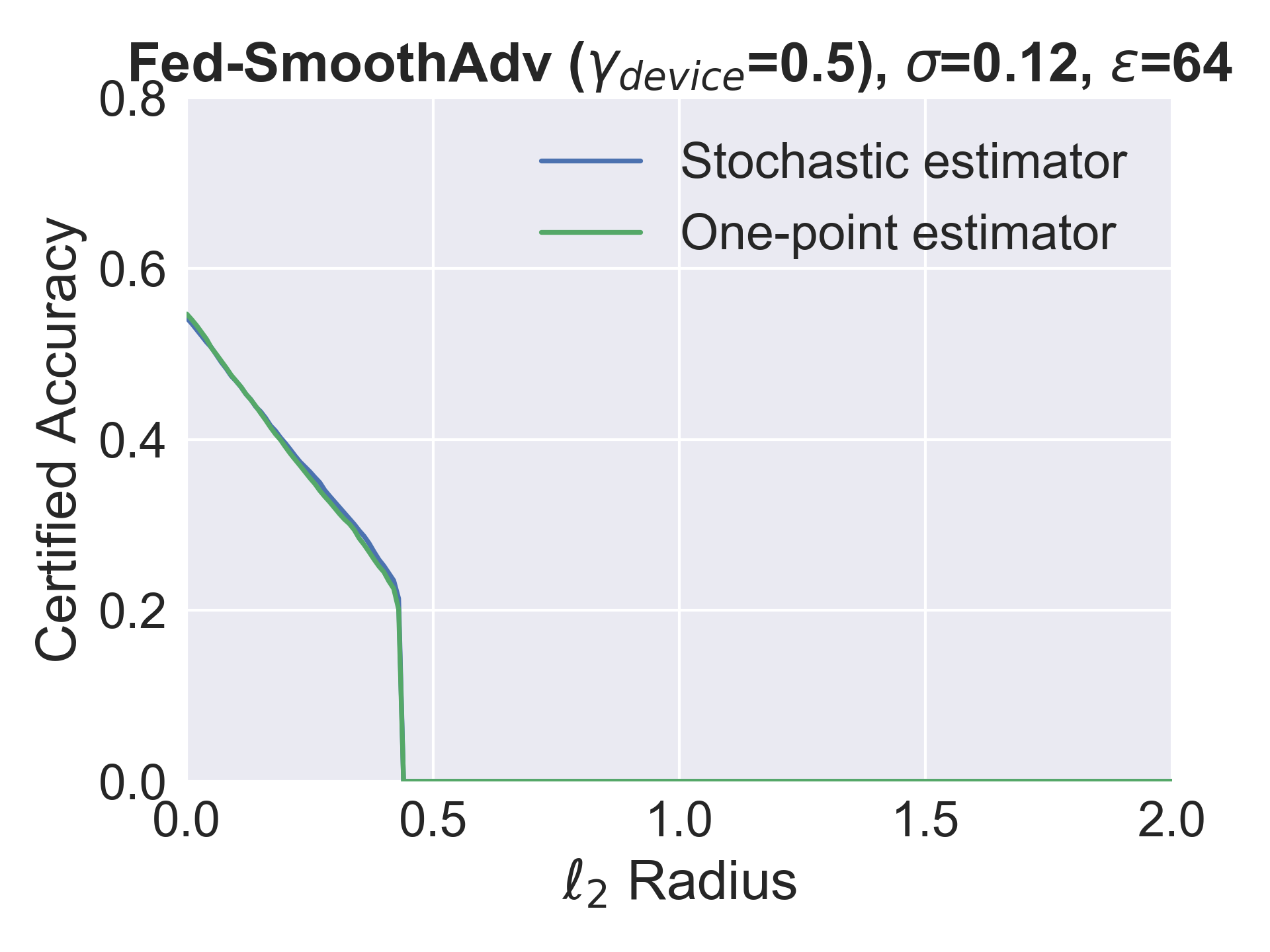}
	\vspace{-4mm}
	\caption{\small{Certified accuracy of {\bf SmoothAdv} and {\bf Fed-SmoothAdv} with $\sigma=0.12$ and $\epsilon=64$.}}\label{fig: c2}
	\vspace{-2mm}
\end{figure}

\begin{figure}[bth]
	\vspace{2mm}
	\centering
	\includegraphics[width=0.32\textwidth,height=0.25\textwidth]{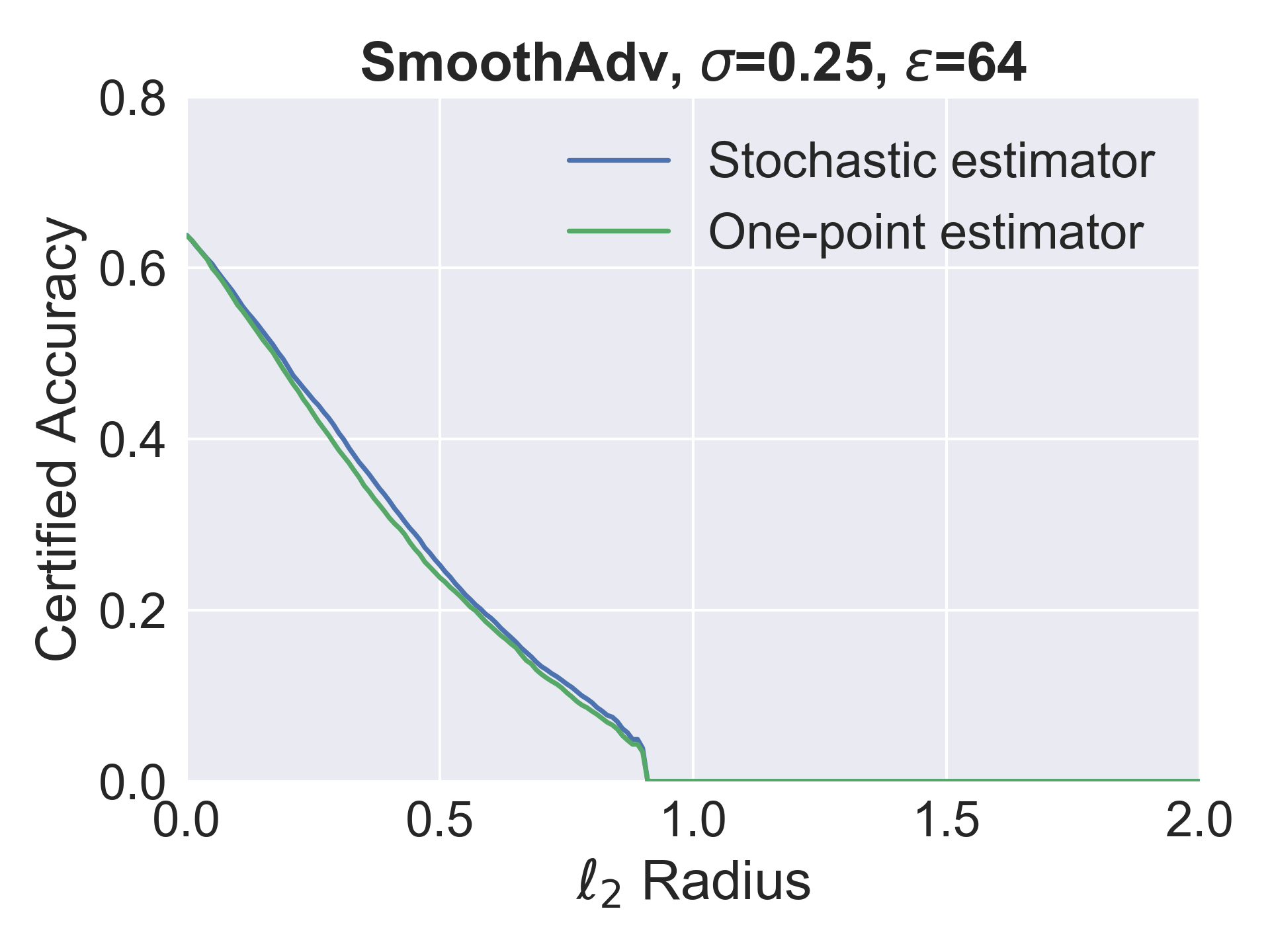}	
	\includegraphics[width=0.32\textwidth,height=0.25\textwidth]{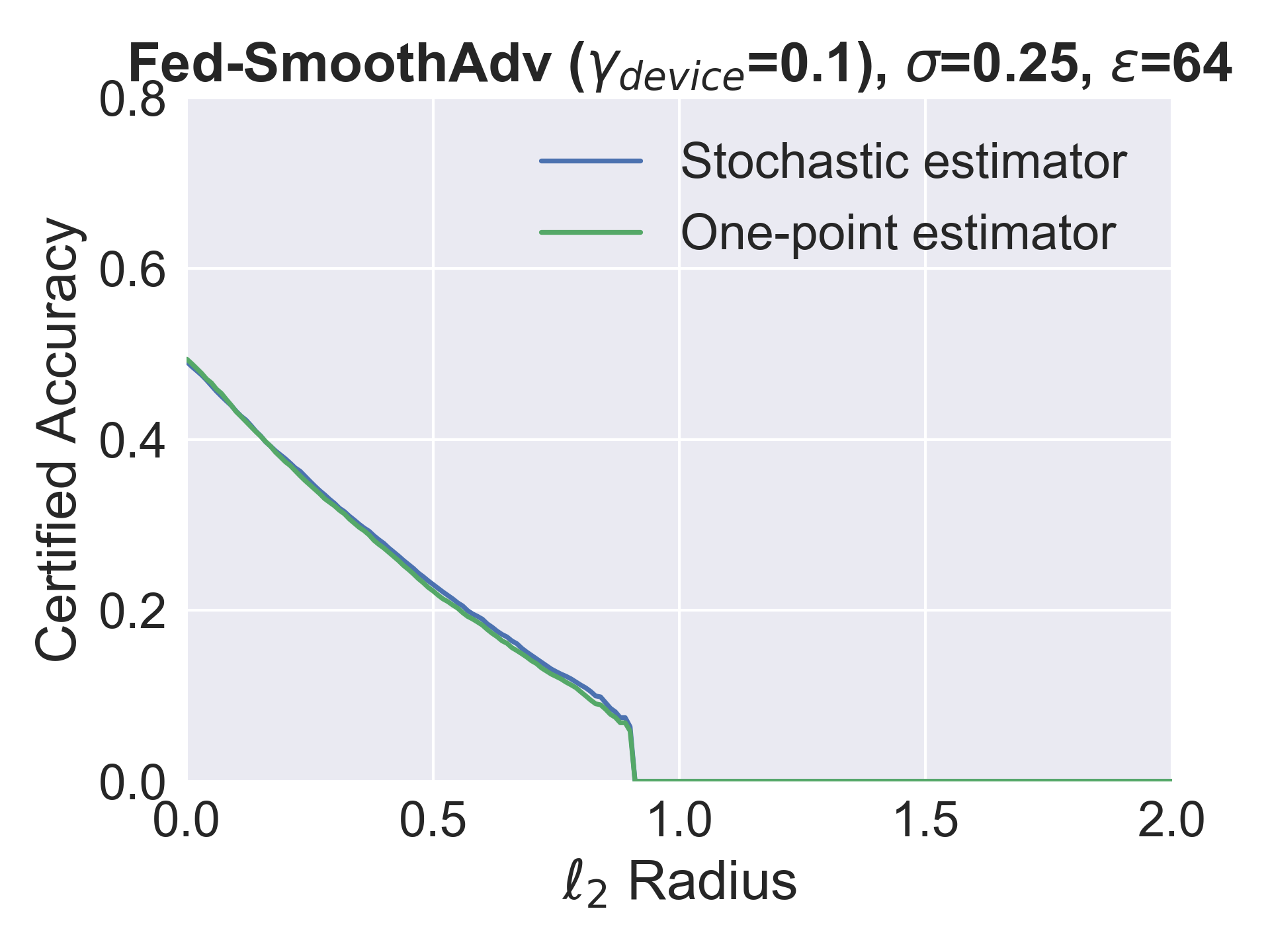}	
	\includegraphics[width=0.32\textwidth,height=0.25\textwidth]{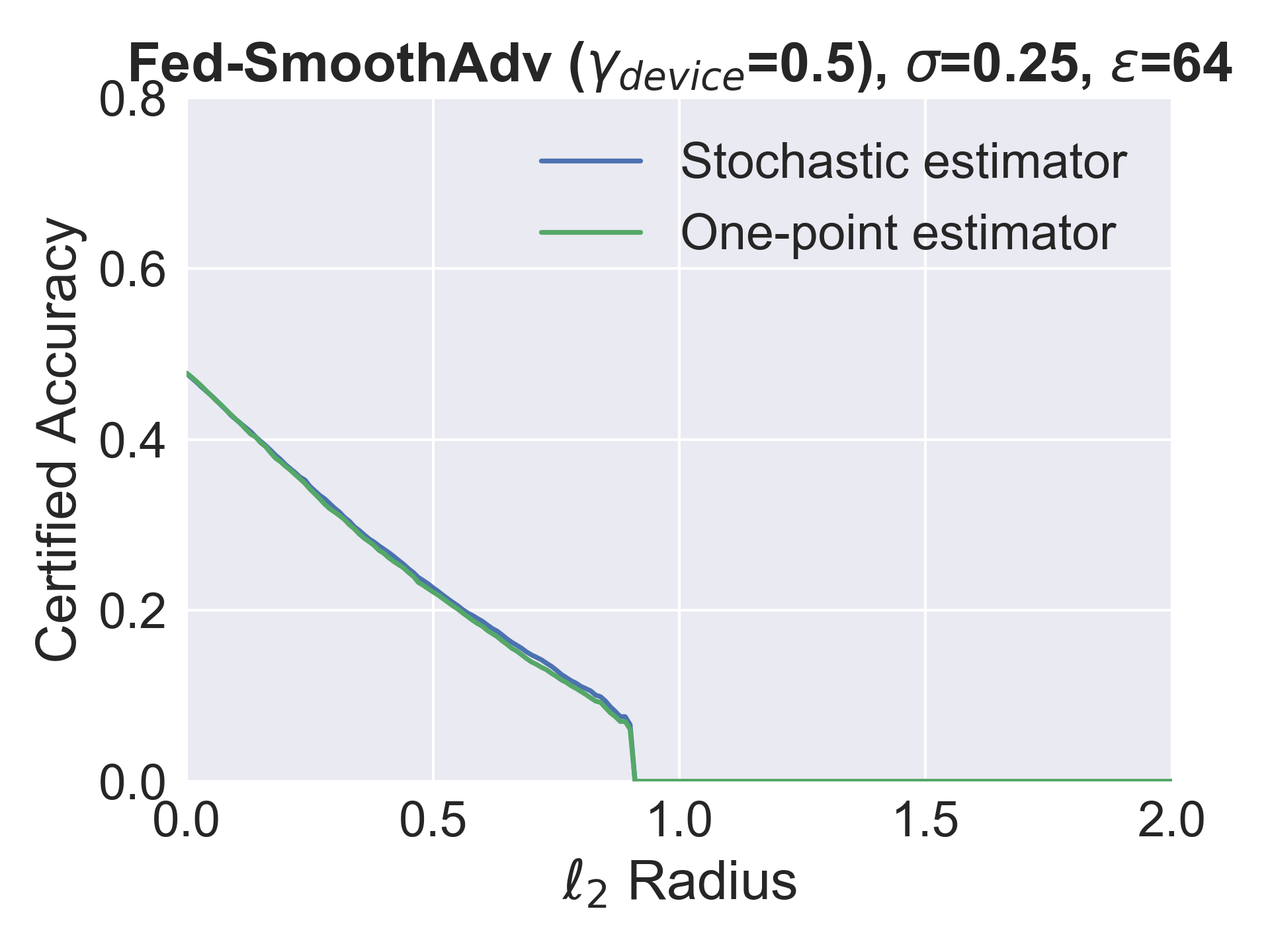}
	\vspace{-4mm}
	\caption{\small{Certified accuracy of {\bf SmoothAdv} and {\bf Fed-SmoothAdv} with $\sigma=0.25$ and $\epsilon=64$.}}\label{fig: c3}
	\vspace{-2mm}
\end{figure}

\begin{figure}[bth]
	\vspace{2mm}
	\centering
	\includegraphics[width=0.32\textwidth,height=0.25\textwidth]{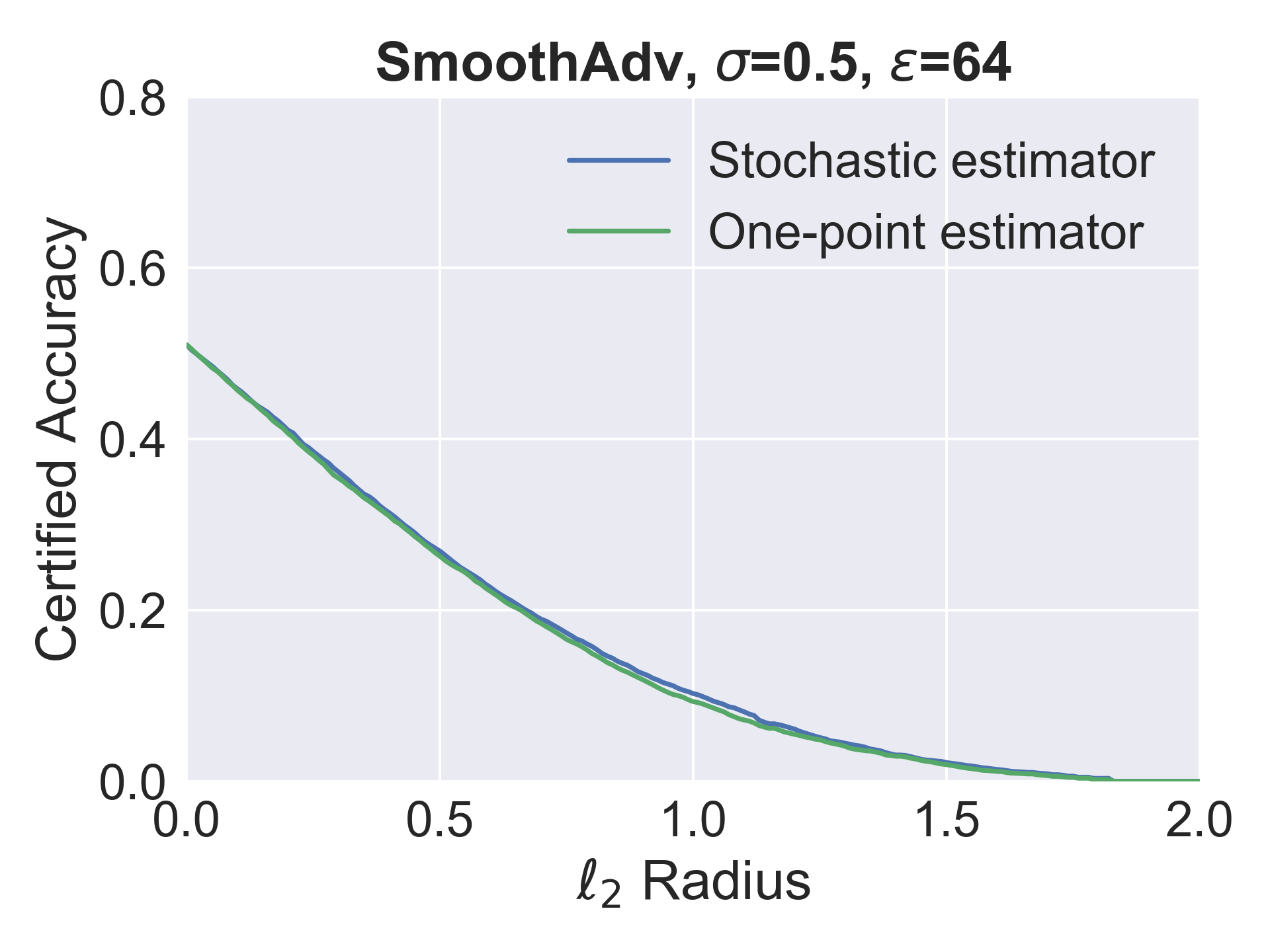}	
	\includegraphics[width=0.32\textwidth,height=0.25\textwidth]{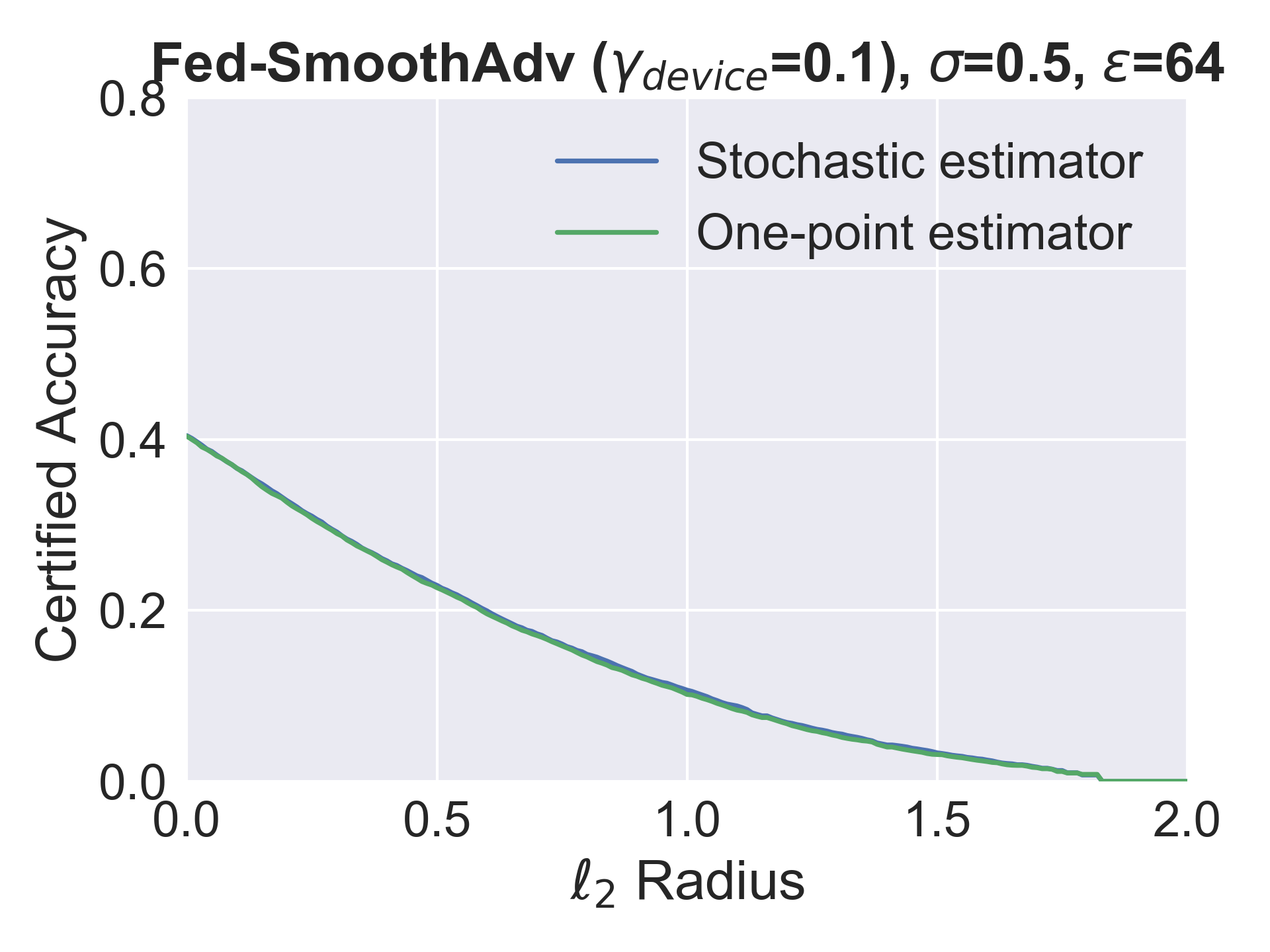}	
	\includegraphics[width=0.32\textwidth,height=0.25\textwidth]{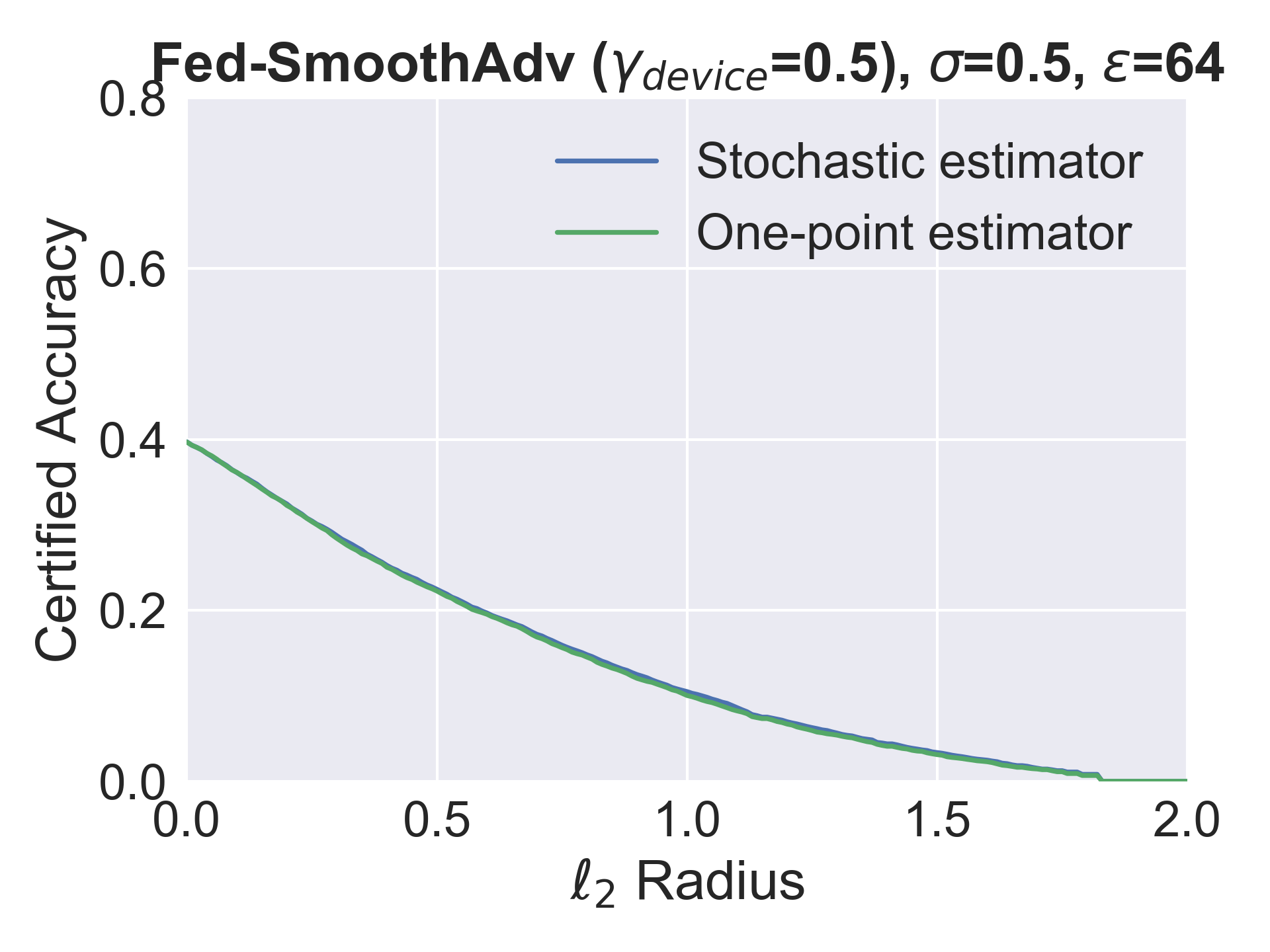}
	\vspace{-4mm}
	\caption{\small{Certified accuracy of {\bf SmoothAdv} and {\bf Fed-SmoothAdv} with $\sigma=0.5$ and $\epsilon=64$.}}\label{fig: c4}
	\vspace{-2mm}
\end{figure}

\begin{figure}[bth]
	\vspace{2mm}
	\centering
	\includegraphics[width=0.32\textwidth,height=0.25\textwidth]{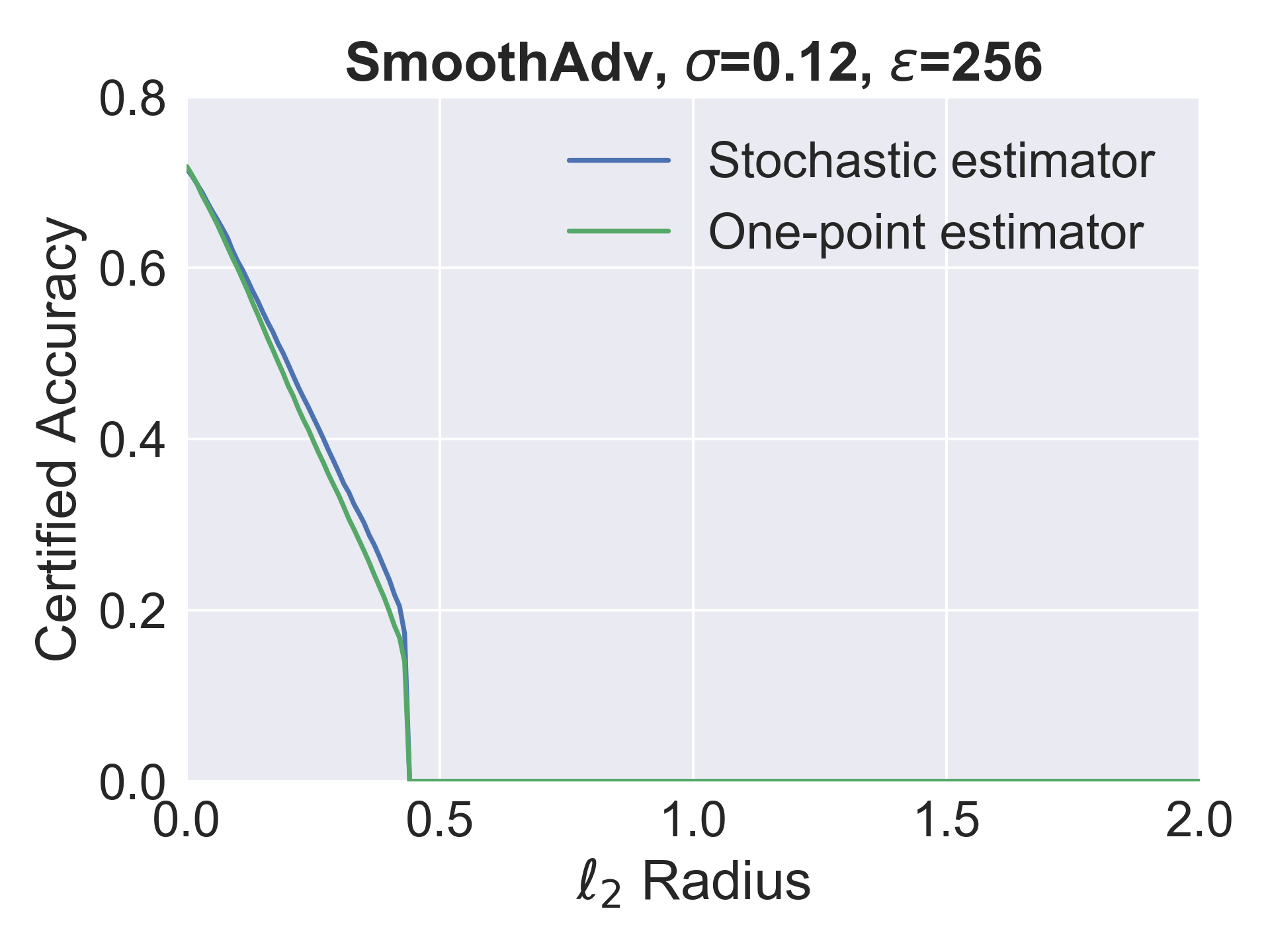}	
	\includegraphics[width=0.32\textwidth,height=0.25\textwidth]{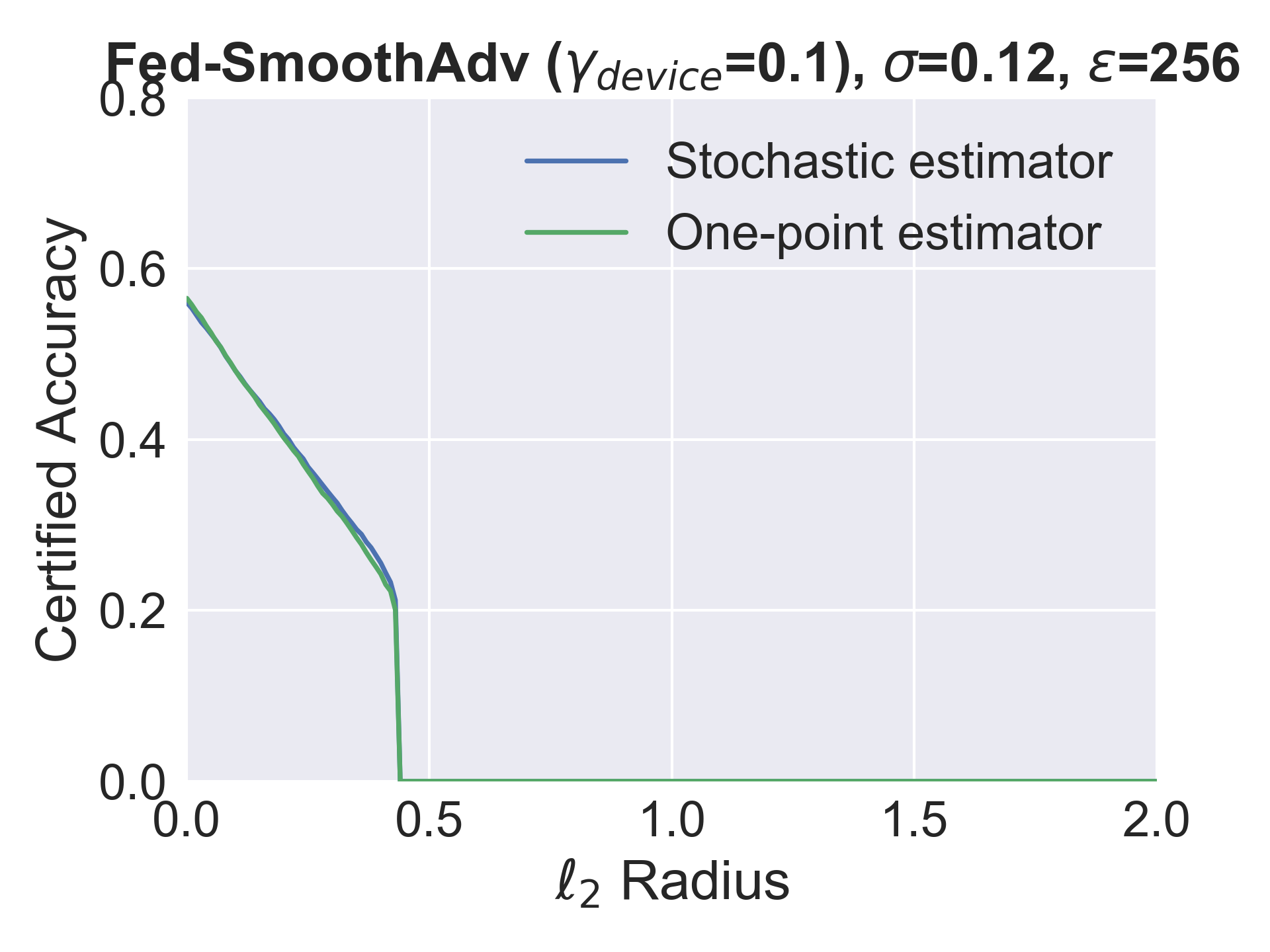}	
	\includegraphics[width=0.32\textwidth,height=0.25\textwidth]{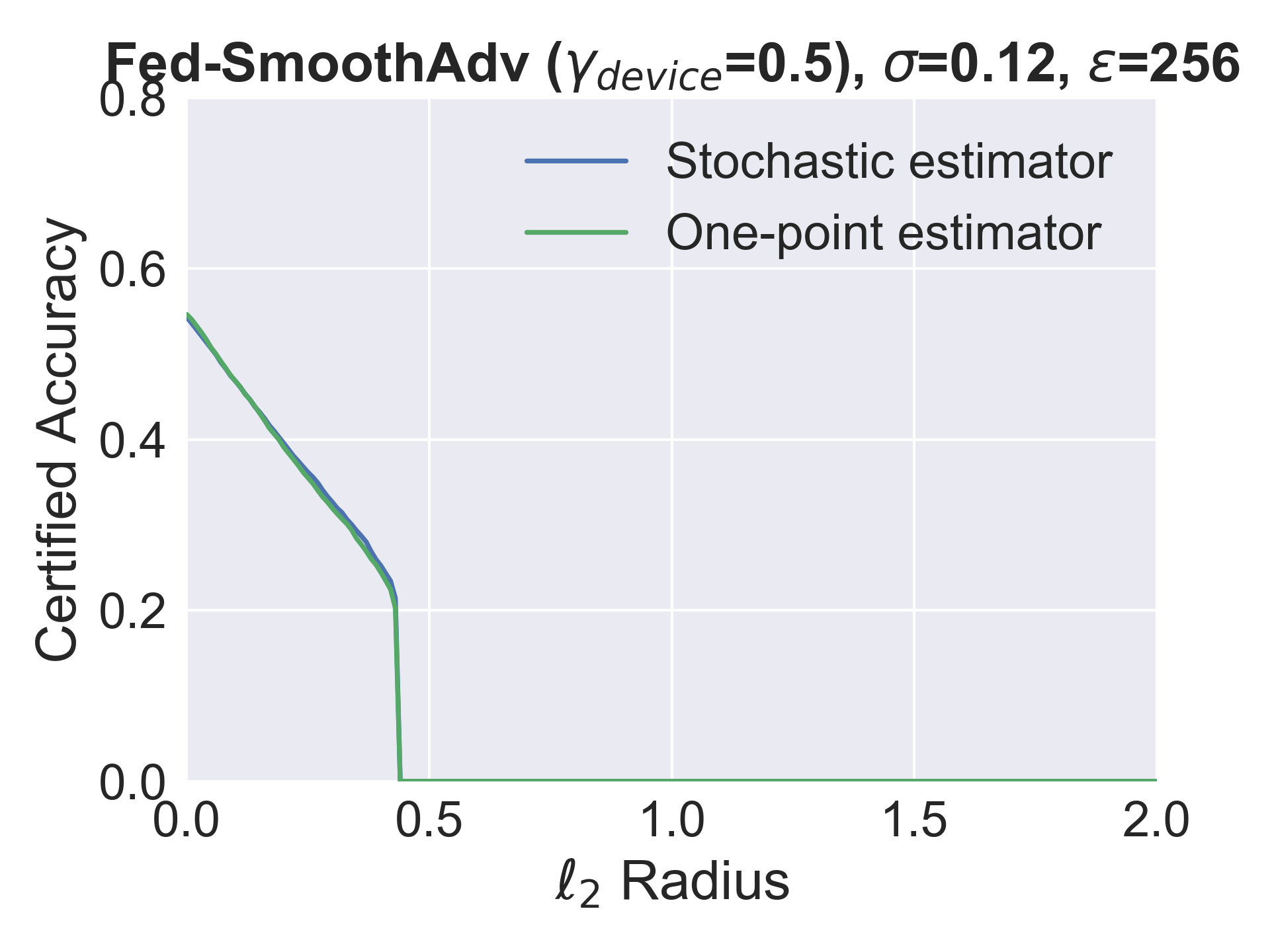}
	\vspace{-4mm}
	\caption{\small{Certified accuracy of {\bf SmoothAdv} and {\bf Fed-SmoothAdv} with $\sigma=0.12$ and $\epsilon=256$.}}\label{fig: c5}
	\vspace{-2mm}
\end{figure}

\begin{figure}[bth]
	\vspace{2mm}
	\centering
	\includegraphics[width=0.32\textwidth,height=0.25\textwidth]{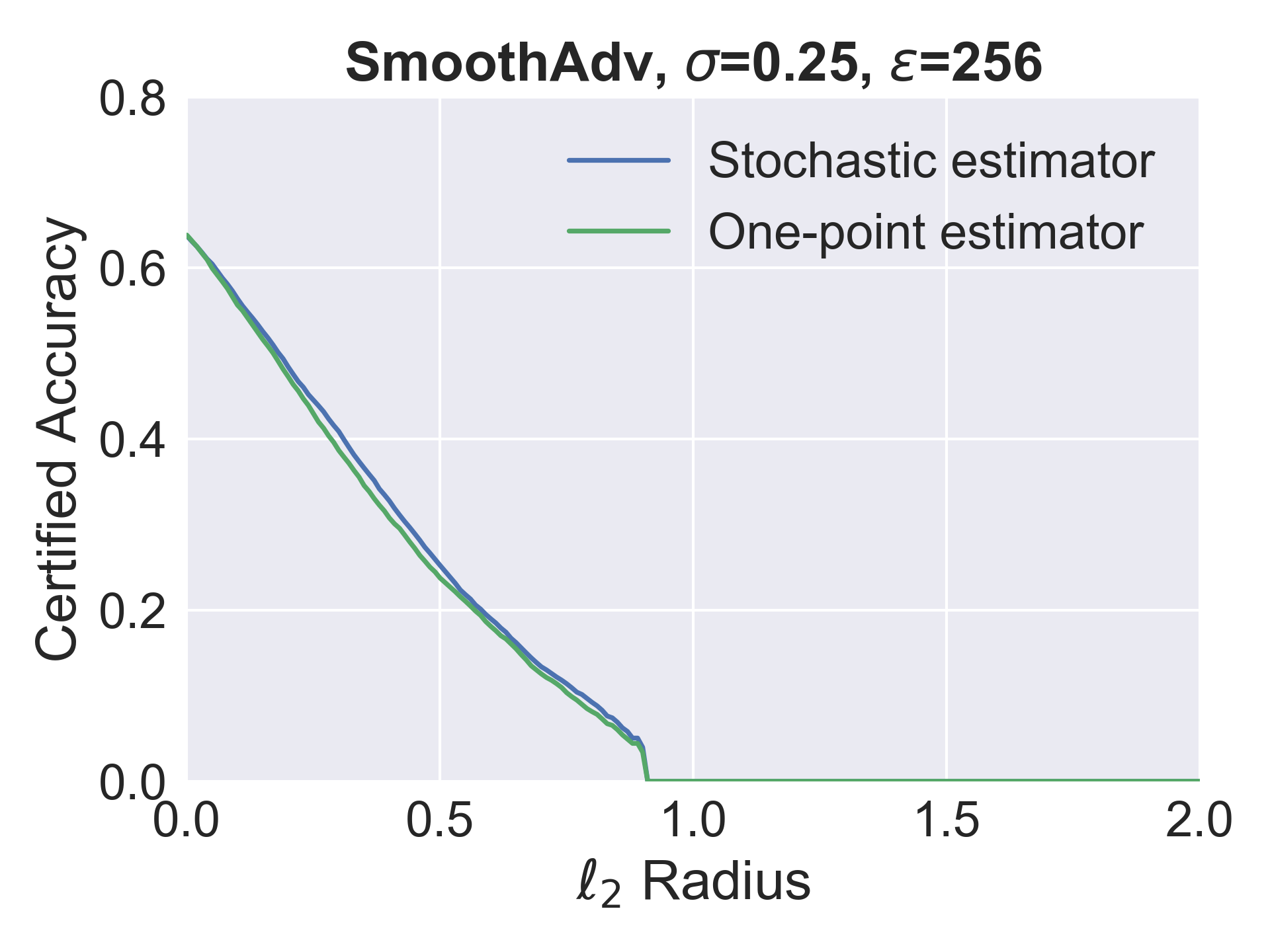}	
	\includegraphics[width=0.32\textwidth,height=0.25\textwidth]{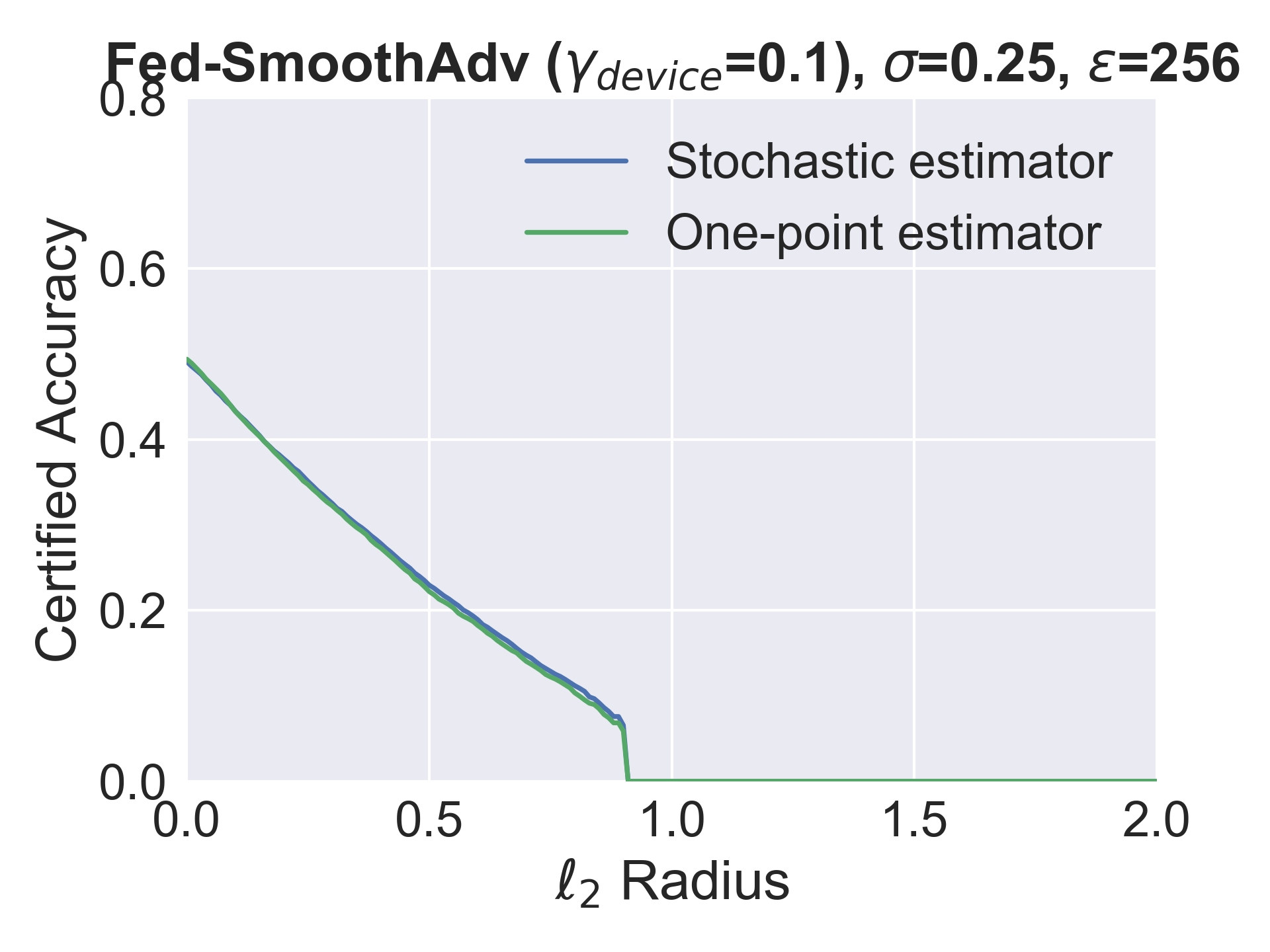}	
	\includegraphics[width=0.32\textwidth,height=0.25\textwidth]{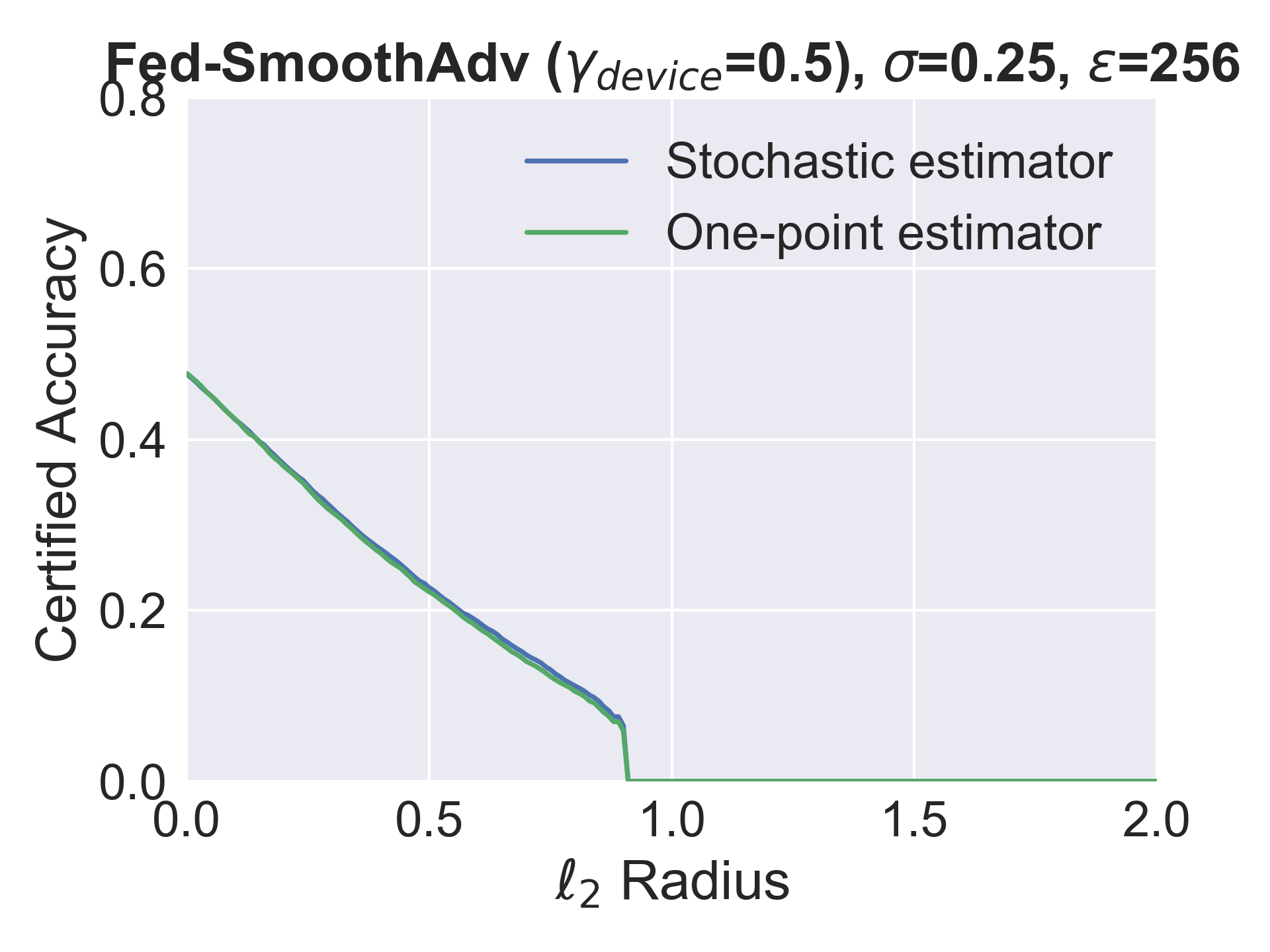}
	\vspace{-4mm}
	\caption{\small{Certified accuracy of {\bf SmoothAdv} and {\bf Fed-SmoothAdv} with $\sigma=0.25$ and $\epsilon=256$.}}\label{fig: c6}
	\vspace{-2mm}
\end{figure}

\begin{figure}[bth]
	\vspace{2mm}
	\centering
	\includegraphics[width=0.32\textwidth,height=0.25\textwidth]{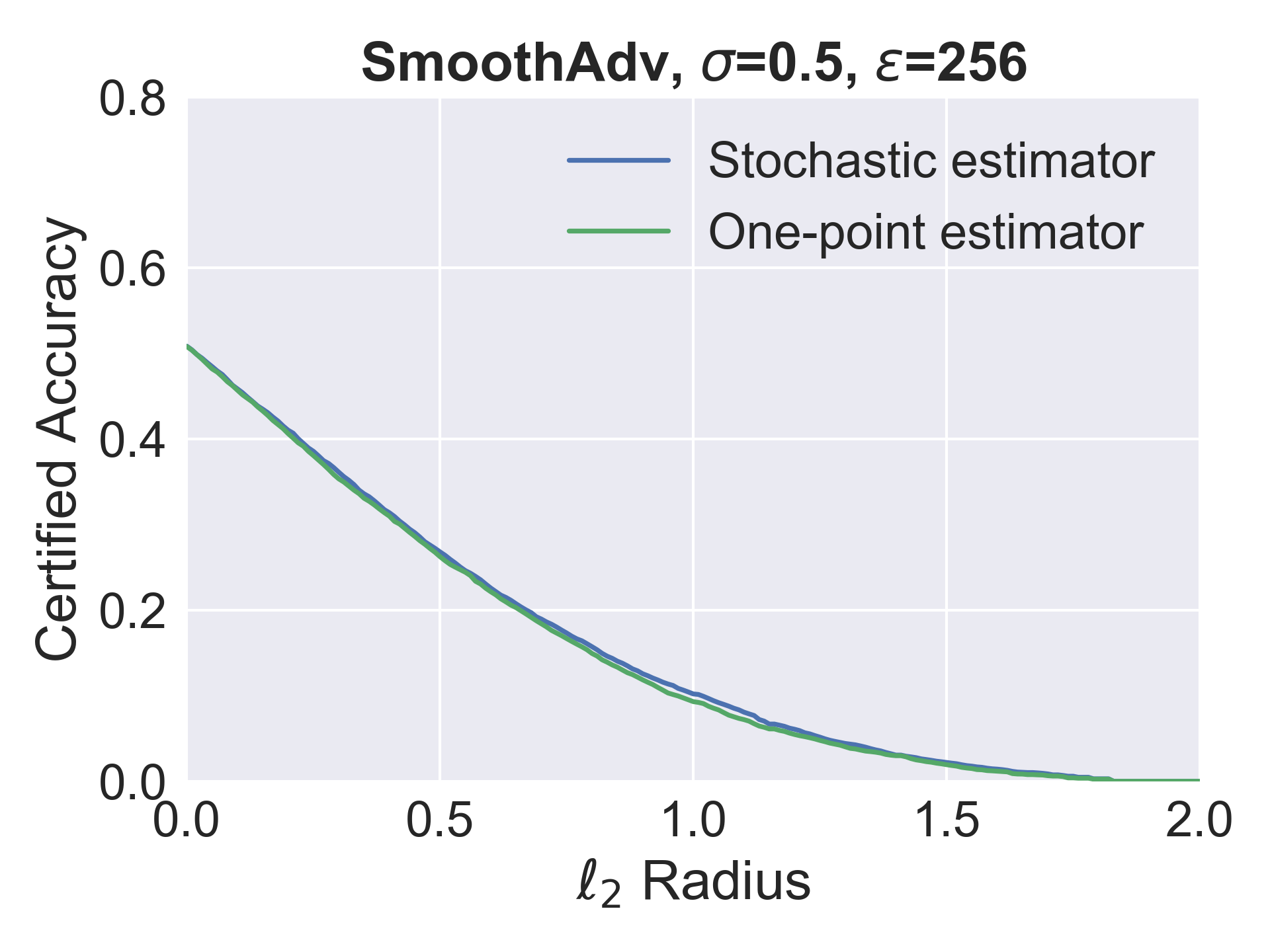}	
	\includegraphics[width=0.32\textwidth,height=0.25\textwidth]{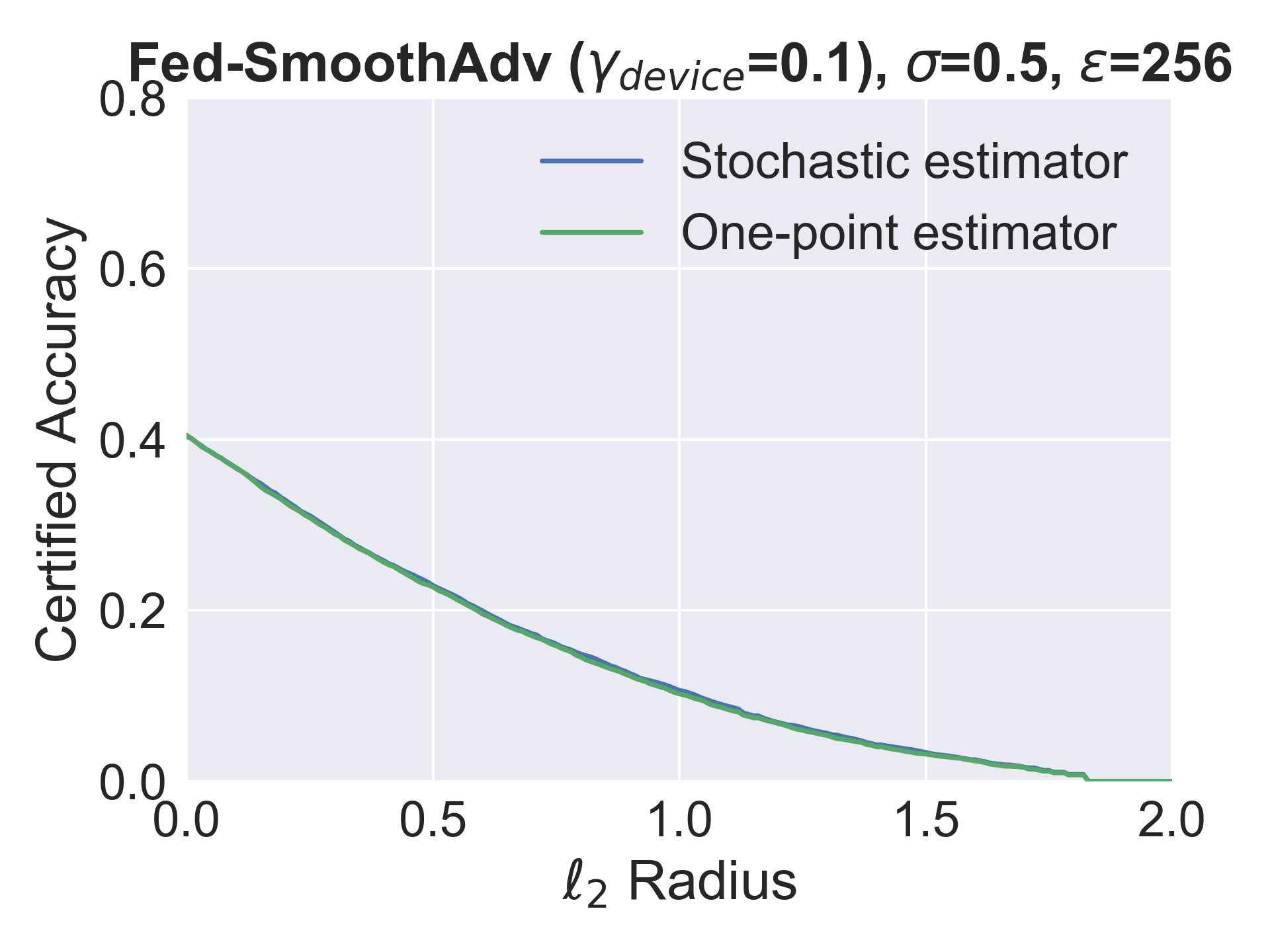}	
	\includegraphics[width=0.32\textwidth,height=0.25\textwidth]{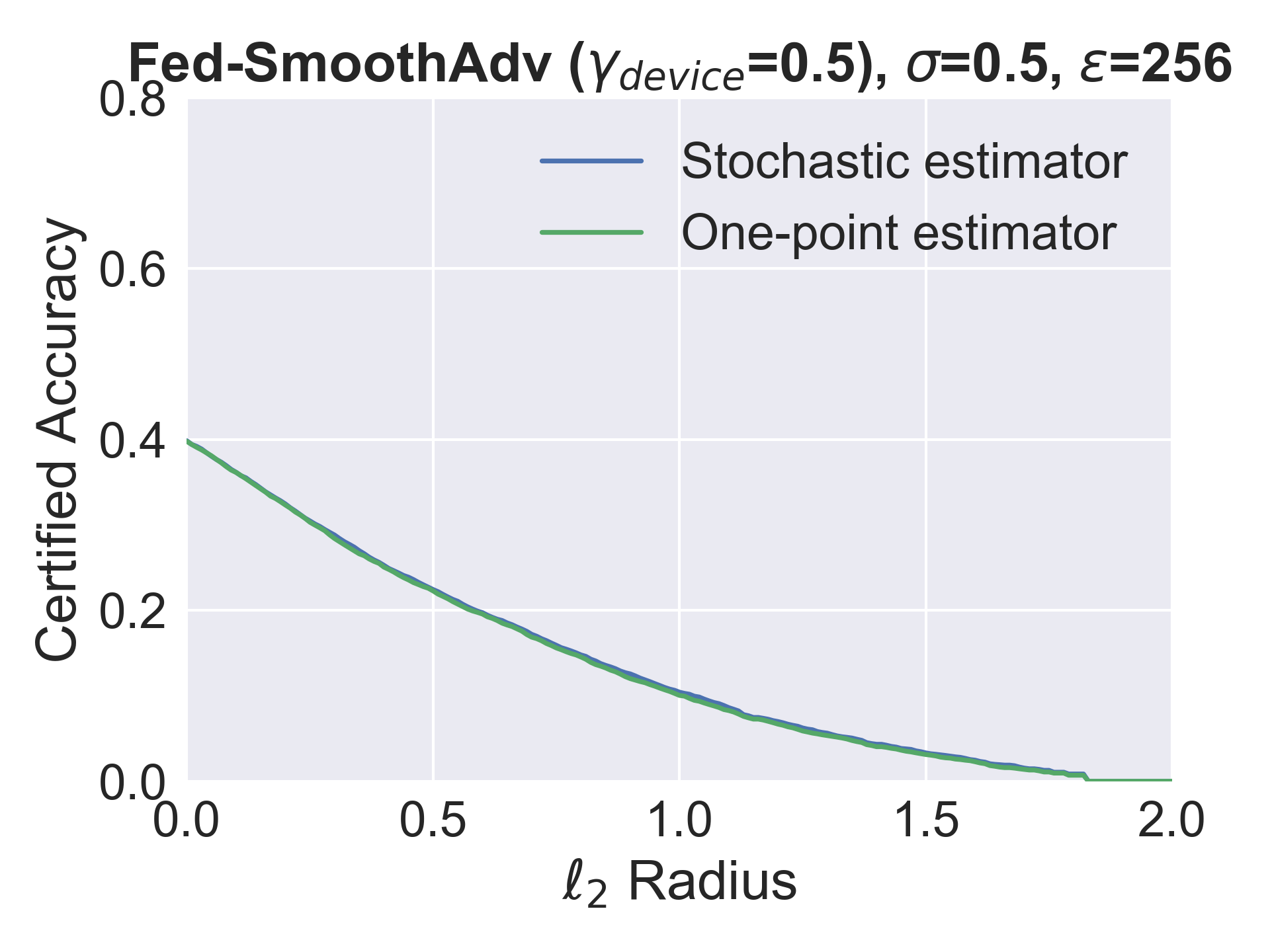}
	\vspace{-4mm}
	\caption{\small{Certified accuracy of {\bf SmoothAdv} and {\bf Fed-SmoothAdv} with $\sigma=0.5$ and $\epsilon=256$.}}\label{fig: c7}
	\vspace{-2mm}
\end{figure}

In this section, we present the certified accuracy results of both {\bf SmoothAdv} and {\bf Fed-SmoothAdv} (with heterogeneity $\gamma_{\text{device}}=0.1, 0.5$) under other choices of $\sigma$ and $\epsilon$. From \Cref{fig: c1}-\Cref{fig: c7}, we observe the same comparison and make the same conclusions as those in \Cref{exp}. 

\end{document}